\documentclass{article}

\usepackage[preprint]{icml2026}

\usepackage{amsmath,amsfonts,bm,amssymb}
\usepackage{mathtools}
\usepackage{amsthm}

\def\eqref#1{equation~\ref{#1}}

\def\1{\bm{1}}

\def\rmD{{\mathbf{D}}}
\def\rmE{{\mathbf{E}}}

\def\rmI{{\mathbf{I}}}

\def\rmS{{\mathbf{S}}}

\def\vc{{\bm{c}}}

\def\vh{{\bm{h}}}

\def\vm{{\bm{m}}}

\def\vp{{\bm{p}}}

\def\mH{{\bm{H}}}

\DeclareMathAlphabet{\mathsfit}{\encodingdefault}{\sfdefault}{m}{sl}
\SetMathAlphabet{\mathsfit}{bold}{\encodingdefault}{\sfdefault}{bx}{n}

\def\gH{{\mathcal{H}}}
\def\gI{{\mathcal{I}}}

\def\sR{{\mathbb{R}}}

\DeclareMathOperator*{\argmin}{arg\,min}

\usepackage{graphicx} 

\usepackage{xcolor}

\usepackage{colortbl}

\usepackage[most]{tcolorbox}

\usepackage{booktabs}

\usepackage{multirow}

\usepackage{adjustbox}

\usepackage{makecell}

\usepackage{tabularx}

\usepackage{longtable}

\usepackage{array}

\usepackage{subcaption}

\usepackage{wrapfig} 

\usepackage[utf8]{inputenc} 

\usepackage[patch=none]{microtype}

\usepackage{stfloats}

\usepackage{lineno}

\usepackage{textcomp}

\usepackage[OT1]{fontenc} 

\usepackage{balance}

\usepackage{enumitem} 

\usepackage{nicefrac}       

\usepackage{amsmath}
\usepackage{amsfonts}
\usepackage{amssymb}
\usepackage{bm}       

\usepackage{amsthm}

\usepackage{pifont}

\usepackage{listings}

\usepackage{verbatim}

\usepackage{hyperref}

\usepackage[textsize=tiny]{todonotes}

\usepackage[capitalize,noabbrev]{cleveref}

\def\HiLi{\leavevmode\rlap{\hbox to \hsize{\color{gray!20}\leaders\hrule height .8\baselineskip depth .5ex\hfill}}}

\definecolor{darkblue}{rgb}{0, 0, 0.5}

\hypersetup{
  colorlinks=true,   
  citecolor=darkblue, 
  linkcolor=darkblue, 
  urlcolor=darkblue   
}

\theoremstyle{plain}

\theoremstyle{definition}

\theoremstyle{remark}

\icmltitlerunning{Large Language Model Reasoning with Implicit Cognition Latent Planning}

\begin{document}

\twocolumn[
  \icmltitle{iCLP: Large Language Model Reasoning with Implicit Cognition Latent Planning}

  \begin{icmlauthorlist}
    \icmlauthor{Sijia Chen}{hkust-gz}
    \icmlauthor{Di Niu}{ualberta}
    
  \end{icmlauthorlist}

  \icmlaffiliation{hkust-gz}{Hong Kong University of Science and Technology (Guangzhou)}
  \icmlaffiliation{ualberta}{University of Alberta}
  
  \icmlcorrespondingauthor{Sijia Chen}{sijiachen@hkust-gz.edu.cn}





  \vskip 0.3in
]

\printAffiliationsAndNotice{}

\begin{abstract}
  Large language models (LLMs), when guided by explicit textual plans, can perform reliable step-by-step reasoning during problem-solving. However, generating accurate and effective textual plans remains challenging due to LLM hallucinations and the high diversity of task-specific questions. To address this, we draw inspiration from human \textit{Implicit Cognition} (\textit{IC}), the subconscious process by which decisions are guided by compact, generalized patterns learned from past experiences without requiring explicit verbalization. We propose \emph{iCLP}, a novel framework that enables LLMs to adaptively generate latent plans (LPs), which are compact encodings of effective reasoning instructions. \emph{iCLP} first distills explicit plans from existing step-by-step reasoning trajectories. It then learns discrete representations of these plans via a vector-quantized autoencoder coupled with a codebook. Finally, by fine-tuning LLMs on paired latent plans and corresponding reasoning steps, the models learn to perform implicit planning during reasoning. Experimental results on mathematical reasoning and code generation tasks demonstrate that, with \emph{iCLP}, LLMs can plan in latent space while reasoning in language space. This approach yields significant improvements in both accuracy and efficiency and, crucially, demonstrates strong cross-domain generalization while preserving the interpretability of chain-of-thought reasoning. The source code is publicly available at \url{https://github.com/AgenticFinLab/latent-planning}.

\end{abstract}


\section{Introduction}
\label{sec:intro}

Large language models (LLMs) employ a step-by-step reasoning process expressed as a chain of thought (CoT) \cite{cot-neurips22} to solve complex problems. Guiding thought generation with explicit plans \cite{ReACT-iclr23}, which are step-wise and question-specific reasoning instructions, is crucial for improving practical usage and reliability of LLMs \cite{llmplansurvey-arxiv24, huang2022inner}. However, preparing such plans effectively for accurate reasoning is inherently challenging \cite{PlanningLLM-neurips23,llmcannotplan-iclr24}, particularly given the high diversity of problems across different tasks.

One preliminary approach to achieving this goal involves prompting LLMs to generate explicit plans using their internal knowledge \cite{ReACT-iclr23,PSPrompting-acl23, AdaPlanner-neurips23}. However, this method is limited by errors in the plans, which arise from the inevitable hallucinations of LLMs. While leveraging external knowledge bases can help mitigate these errors \cite{RPG-emnlp24, KnowAgent-naacl24}, accessing useful information from them is time-consuming, and many tasks lack effective knowledge bases altogether. More promising recent efforts \cite{ReACT-iclr23,LearnPlanningReasoning-emnlp24,AutoAct-acl24,PlaSma-iclr24} focus on fine-tuning LLMs on automatically or manually synthesized samples with explicit plans. Unfortunately, LLMs fine-tuned in this manner still struggle to achieve better performance because the plans required by problems within a single task, as well as across tasks, are vast in number and highly diverse. 

We argue that these mechanisms do not align with human wisdom, known as \textit{Implicit Cognition} (\textit{IC}) \cite{ImplicitCognition}, through which we learn from experiences to summarize implicit patterns that shape our subconscious mind \cite{subconscious}, allowing us to use these patterns to solve new problems without explicitly verbalizing them. These patterns contain abstract rules that reflect high-level common knowledge, capable of generalizing across different problems. Additionally, our preliminary experiments on visualizing the representations of explicit plans distilled from CoTs of different questions reveal clear clustering for plans from the same task, along with a certain level of overlap indicating their reuse.

Therefore, this paper mimics IC by proposing a framework called \emph{iCLP}, which enables any LLM to generate latent plans (LPs) in a hidden space, effectively guiding step-by-step reasoning in the language space. Our key insight for its effectiveness is that LPs, built upon summarizing experiences, are analogous to the subconscious mind in humans. Similar to how the subconscious mind serves as a flexible and adaptive guidance system in the brain for diverse problems \cite{SubconsciousPower}, LPs, due to their commonality and reusability for reasoning guidance, are small in scale, making them generalizable across tasks.

To construct this space, \emph{iCLP} first prompts an off-the-shelf LLM to summarize explicit plans from a collection of effective CoT traces. Subsequently, \emph{iCLP} borrows the encoding module of ICAE \cite{icae-iclr24} to map these distilled plans into a small set of memory slots that compress their semantics; it then derives generic slot representations from a codebook learned via a vector-quantized autoencoder \cite{vqgan-cvpr21}, trained end-to-end with plan reconstruction. By treating codebook indices as special tokens for the LLM, we directly obtain the \textit{latent plans}, which serve as compact encodings of the plans within the learned codebook. Finally, by integrating them into the original samples, we reformulate each sample into the form: ($user$: question, $assistant$: latent plans and CoTs). Fine-tuning any LLM on these samples enables the model to internalize the intelligence of \textit{IC}, empowering it to perform latent planning for reliable, step-wise reasoning.

We conduct evaluations on mathematical reasoning and code generation tasks. For accuracy, supervised fine-tuning of small LLMs, such as Qwen2.5-7B, with \emph{iCLP} on datasets like \texttt{MATH} and \texttt{CodeAlpaca} yields substantial gains, achieving performance competitive with GRPO \cite{GRPO}, which relies on reinforcement learning. For efficiency, LLMs enhanced with \emph{iCLP} reduce token cost by 10\% on average compared to zero-shot CoT prompting. For generality, cross-dataset evaluations show that fine-tuned models applied to \texttt{AIME2024} and \texttt{MATH-500} achieve more than a 10\% average accuracy improvement over base models. Similarly, on \texttt{HumanEval} and \texttt{MBPP}, we observe a 9\% gain. Moreover, LLMs fine-tuned with \emph{iCLP} outperform all baselines, including those trained with long CoT samples and latent CoT reasoning, while maintaining interpretability.

\section{Related Work}
\label{sec:related}

Large language models (LLMs) have the capability to solve problems using \textbf{step-by-step reasoning} \cite{llmszeroshot-nips22}, where each step addresses a sub-problem and is described textually as a thought, thereby collectively forming a Chain of Thought (CoT) \cite{cot-neurips22}. However, inherent hallucinations in LLMs can lead to ineffective thoughts. To address this issue, prompting methods have been proposed to extend CoT reasoning by incorporating additional searching \cite{self-consistency-iclr22, least2most-iclr23, complexity-iclr23, tot-nips23} or self-reflection mechanisms \cite{bot-iclr24, tr-icml24, selfcheck-iclr24}.

Instead of relying on resource-intensive approaches, \textbf{enhancing reasoning with step-wise planning} aims to directly guide LLMs using a plan --- a textual instruction that specifies \textit{what to do} at each step for reliable reasoning \cite{llmplansurvey-arxiv24, huang2022inner}. Prior works \cite{PSPrompting-acl23, AdaPlanner-neurips23, np-nips23} Prompt LLMs to generate question-specific plans or premises to guide step-wise reasoning. 
Building on this idea, subsequent works \cite{ReACT-iclr23, RPG-emnlp24, EnPL-emnlp24, llmmcts-nips23, agentplanning-nips24} propose to guide LLMs by synthesizing new reasoning processes in the form of planning trajectories, where each step begins with a plan and is followed by the generation of a corresponding solution step. However, as it remains difficult for LLMs to produce effective and coherent plans \cite{PlanningLLM-neurips23, llmcannotplan-iclr24, BarriersPlanning-acl25}, recent methods \cite{ ReACT-iclr23, LearnPlanningReasoning-emnlp24,AutoAct-acl24, PlaSma-iclr24, agentplanning-nips24} fine-tune LLMs using synthesizes planning trajectories. 
However, the plans used in these approaches are still tightly coupled with specific questions and task contexts. As a result, they often lack generalizability and are susceptible to errors in detailed content. In contrast, our work focuses on latent plans, which provide high-level, concise, and generalizable instructions.

\textbf{Reasoning in a latent continuous space} has been increasingly explored in recent research \cite{LaRS-emnlp24, Coconut-arxiv24, ByteLatent-arxiv24, TokenAssorted-arxiv25,LCM-arxiv24,ContinuousConcepts-arxiv25}, demonstrating promising improvements in both accuracy and efficiency. Similar to our approach of learning latent plans for CoT rationales, LaRS \cite{LaRS-emnlp24} proposes constructing a latent space of rationales via unsupervised learning, enabling LLMs to retrieve latent rationales for given questions. 
The TokenAssorted in \cite{TokenAssorted-arxiv25} abstracts away initial reasoning steps using latent discrete tokens generated by VQ-VAE \cite{vqvae-neurips17}. CoCoMix \cite{ContinuousConcepts-arxiv25}, closely related to our work, combines discrete next-token prediction with continuous concept representations learned via a pretrained sparse autoencoder. 
Although the ideas presented in these methods are closely aligned with ours in blending latent representations with language tokens during reasoning, our work makes a significant advance by explicitly separating the planning and reasoning phases: planning is performed in a latent space, while reasoning is carried out in natural language.

\section{Preliminary and Motivation}
\label{sec:preli}

\subsection{Step-wise Reasoning with Planning}
\label{subsec:planning4reason}

Given a question \( Q \), the large language model (LLM) denoted as \( f \) with parameters \( \bm{\theta} \), generates a chain of thoughts (CoTs), denoted as \( \vc_{1\dots n} = \left[\vc_1, \vc_2, \dots, \vc_n\right] \), where each \( \vc_i \) with \( i \in [1,\dots,n] \) is a textual description of the thought at the \( i \)-th reasoning step. Each thought derives formally from conditional sampling \(\vc_{i} \sim f_{\bm{\theta}}\left(\vc_{i}|Q, \vc_{1\dots i-1}\right)\). The target of \(n\) steps reasoning is to produce the predicted solution \( \widetilde{y} \) in \( \vc_n \) to match the ground truth \( y \). With \textit{explicit plans} as shown in Figure~\ref{fig:main-structure}, denoted as \(\bm{p}_{1 \dots n} = \left[\bm{p}_1, \bm{p}_2, \dots, \bm{p}_n\right]\), where each represents a textual instruction outlining \textit{what to do} in a single reasoning step, the LLM can gain prior knowledge on how to organize \(\vc_{1\dots n}\) for reliable problem-solving. Thus, we reformulate CoTs as $\vc_{1\dots n} = \left[\vc_1, \vc_2, \dots, \vc_n \mid \bm{p}_{1 \dots n}\right]$, meaning that the \(i\)-th thought is represented as \(\vc_{i} \sim f_{\bm{\theta}}\left(\vc_{i} \mid Q, \vc_{1 \dots i-1}, \bm{p}_{1 \dots i-1}, \bm{p}_i\right)\). The guidance of \(\bm{p}_i\) reduce the randomness and uncertainty of LLMs in generating the thought.

To enable reasoning with plan guidance, a direct prompting approach, \emph{PS}, performs sampling via \(\bm{p}_{1 \dots n} \sim f_{\bm{\theta}} \left(\bm{p}_{1 \dots n} \mid \rmI\right)\), where $\rmI$ represents a customized prompt. Other approaches, such as Trajectories Synthesis, REACT \cite{ReACT-iclr23},  AUTOACT \cite{AutoAct-acl24}, and PlaSma \cite{PlaSma-iclr24}, formulate the sample as that each reasoning step \(\vc_{i}\) condition on \(\bm{p}_{i} \sim f_{\bm{\theta}} \left(\bm{p}_{i} \mid Q, \vc_{1 \dots i-1}, \bm{p}_{1 \dots i-1}, \rmI\right)\), allowing the LLM to be fine-tuned to first plan then reason.

\subsection{Explicit Plans present Commonality across Problems}
\label{subsec:motivation}

\begin{figure*}[t]
    \centering
    
    \begin{subfigure}[b]{0.19\textwidth}
        \includegraphics[width=\linewidth]{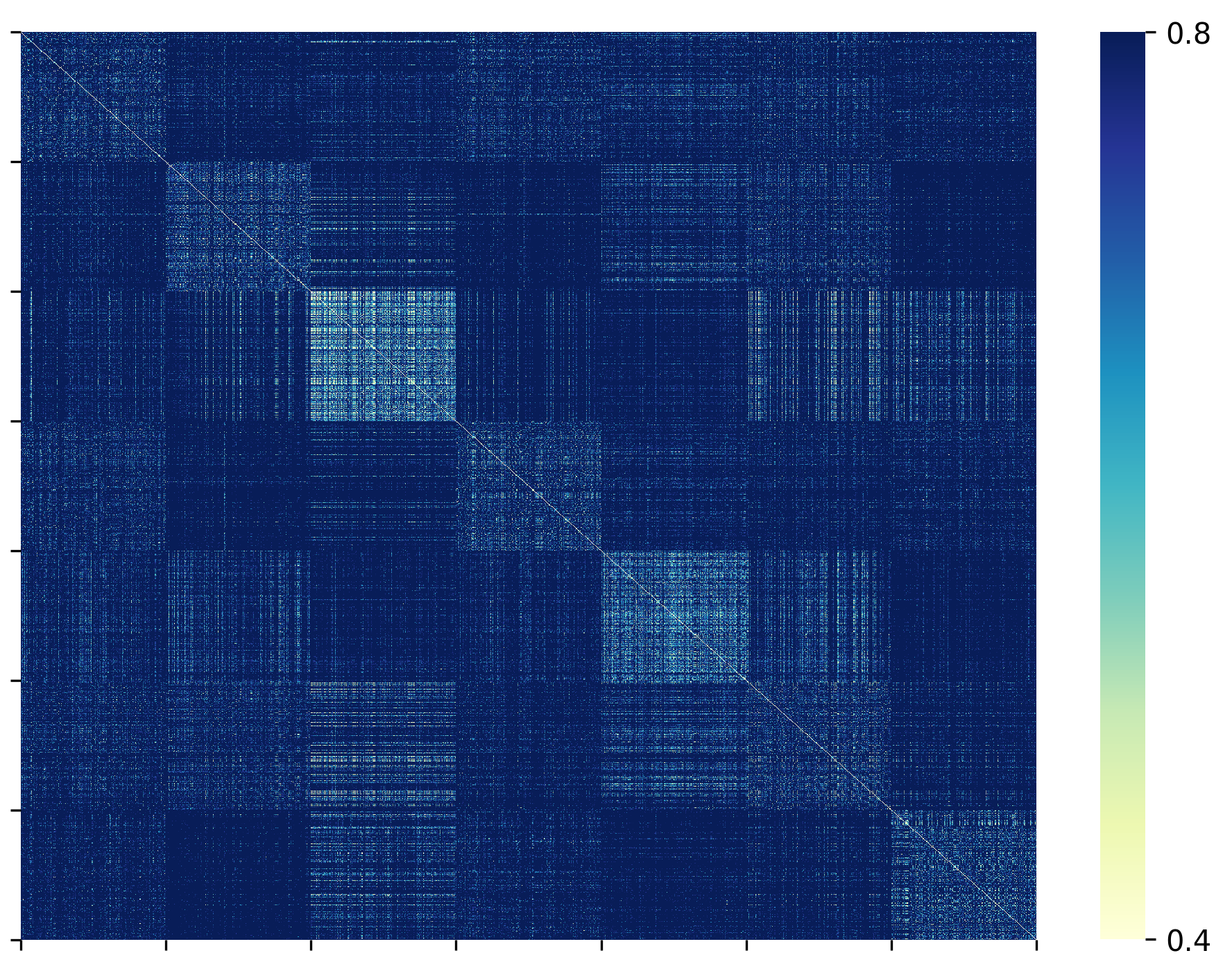} 
        \caption{}
        \label{fig:question}
    \end{subfigure}
    \hfill
    \begin{subfigure}[b]{0.19\textwidth}
        \includegraphics[width=\linewidth]{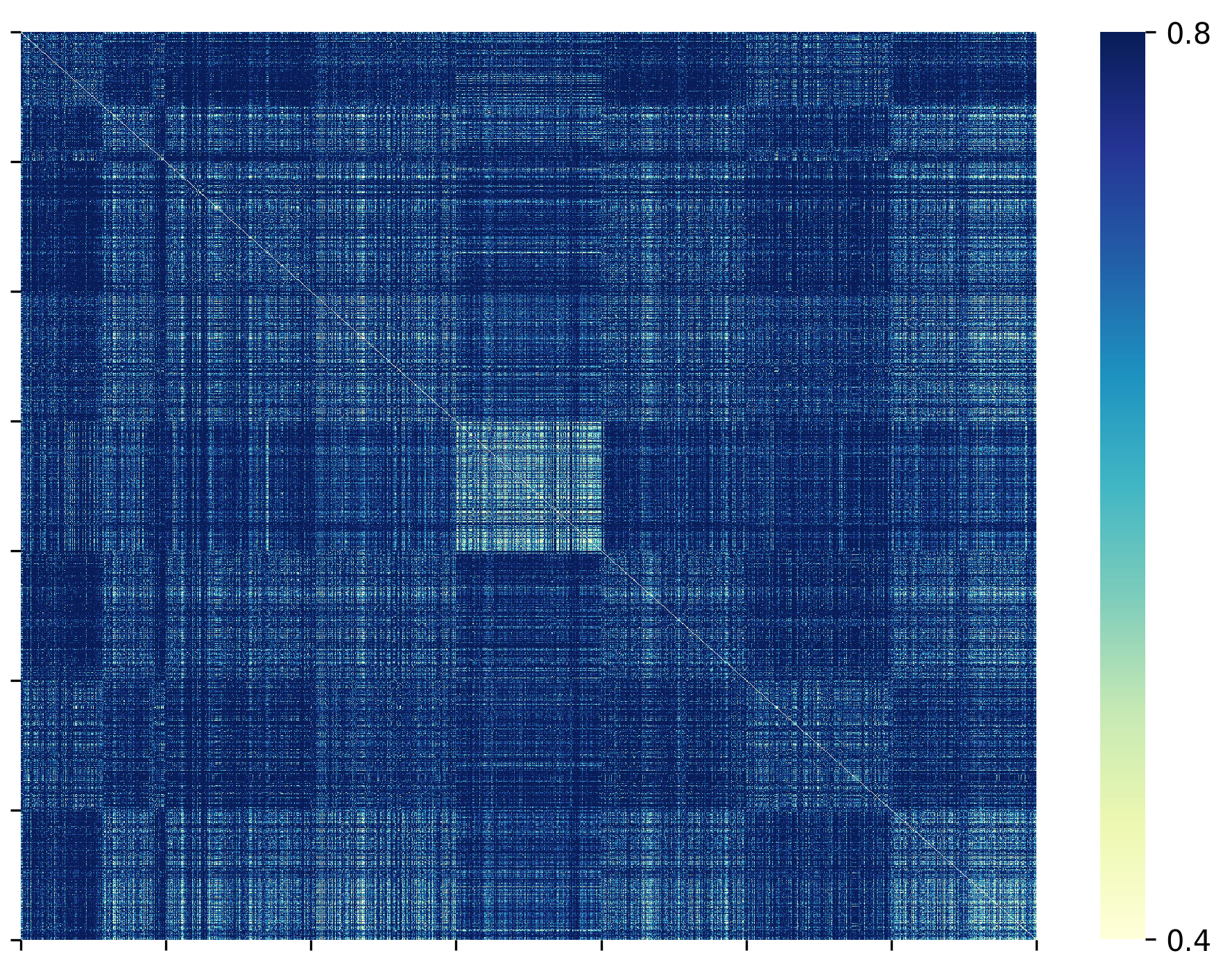} 
        \caption{}
        \label{fig:explicitplan1}
    \end{subfigure}
    \hfill
    \begin{subfigure}[b]{0.19\textwidth}
        \includegraphics[width=\linewidth]{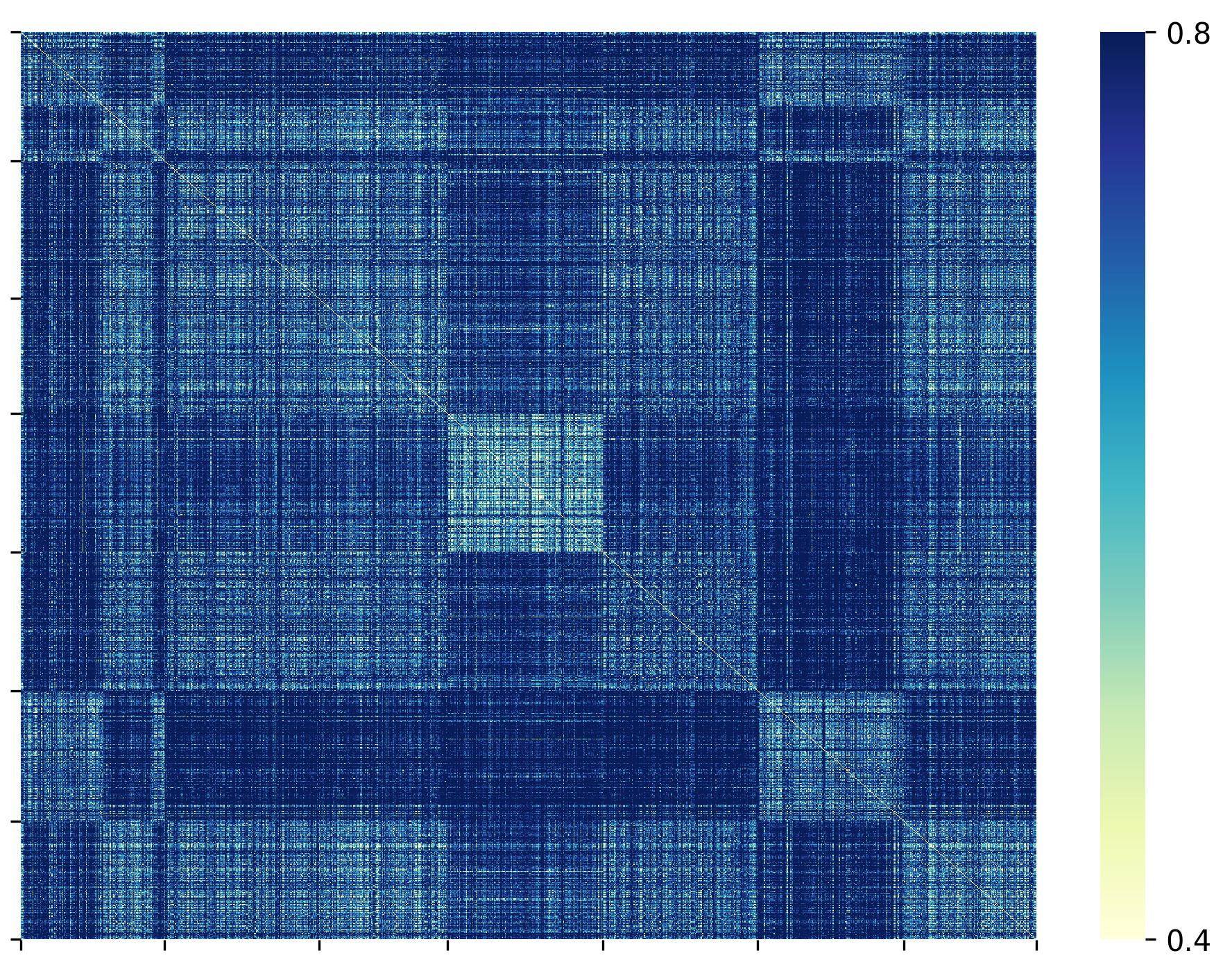} 
        \caption{}
        \label{fig:explicitplan3}
    \end{subfigure}
    \hfill
    \begin{subfigure}[b]{0.19\textwidth}
        \includegraphics[width=\linewidth]{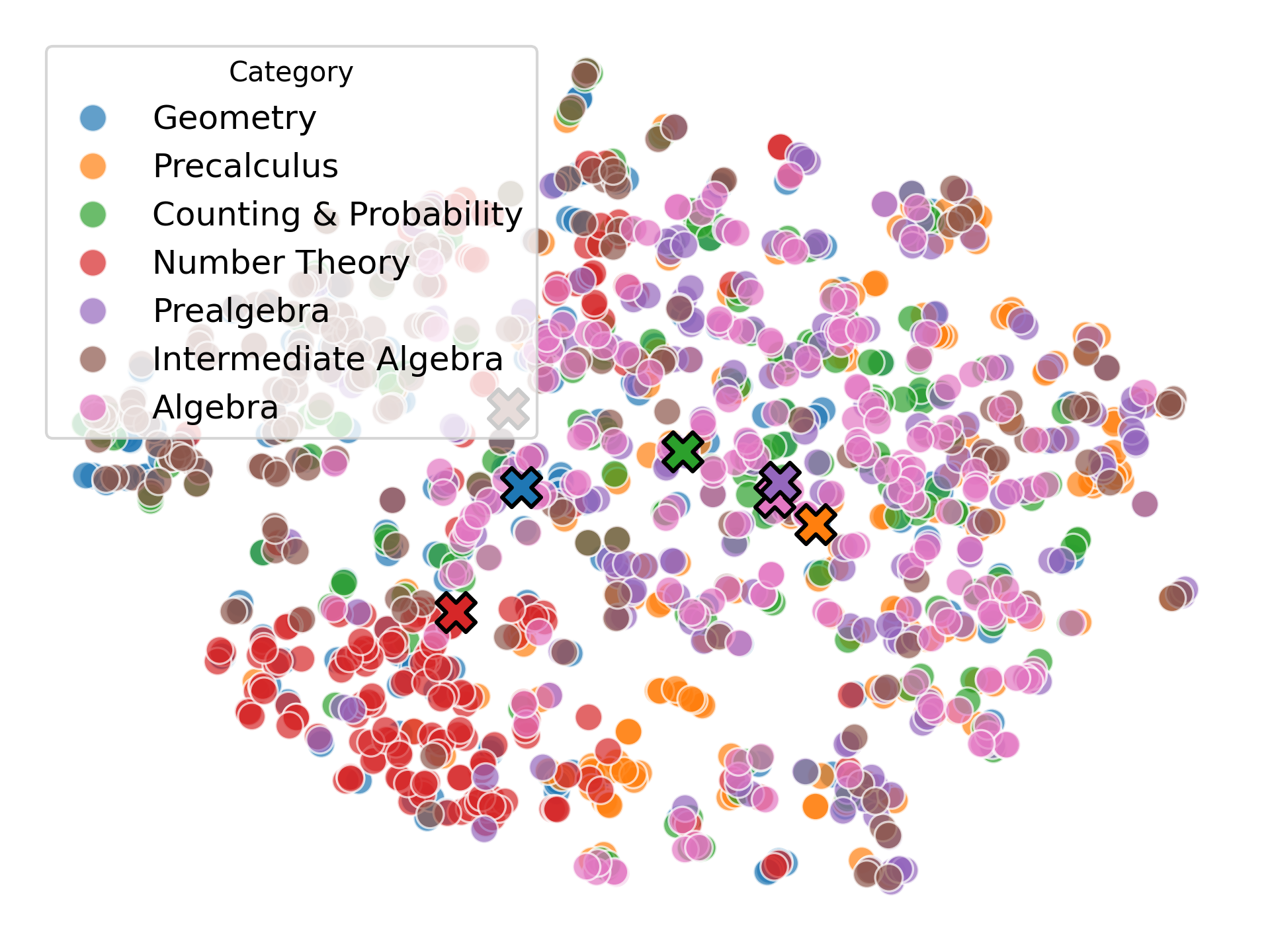} 
        \caption{}
        \label{fig:explicitplan1cluster}
    \end{subfigure}
    \hfill
    \begin{subfigure}[b]{0.19\textwidth}
        \includegraphics[width=\linewidth]{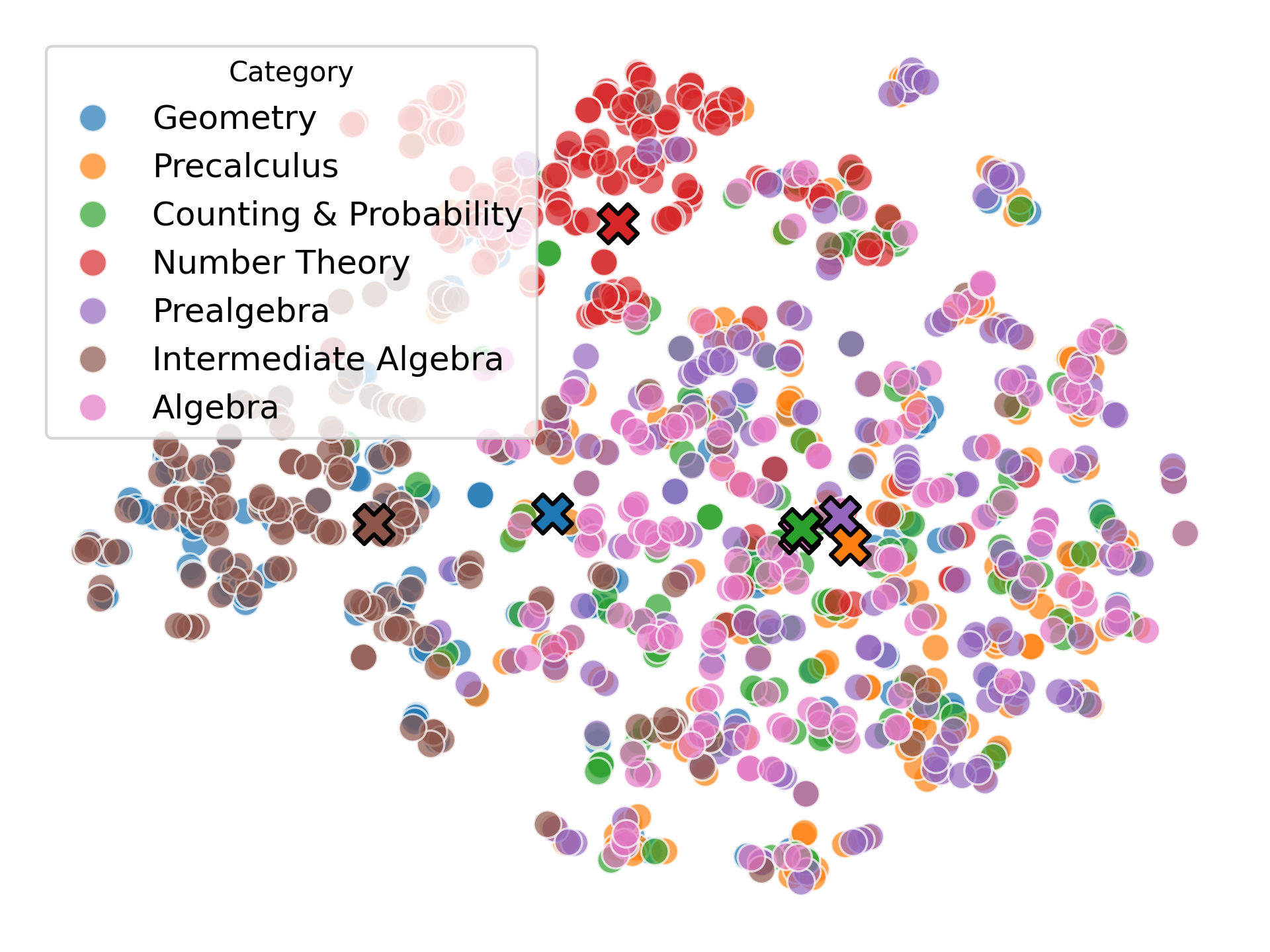} 
        \caption{}
        \label{fig:explicitplan3cluster}
    \end{subfigure}
    \caption{Illustration of relations between explicit plans of questions from different categories. We extract \(200\) samples from each category of the \texttt{MATH} dataset's 7 categories and prompt the LLM to decompose the answers into individual steps, followed by summarizing their explicit plans. (a)(b)(c) display the encoding distances between pairs of items: questions, explicit plans of Step 1 and Step 3, respectively. (d)(e) show the encoding clusters of the explicit plans of Step 1 and Step 3.}
    \label{fig:comparsion}
    \vspace{-4mm}
\end{figure*}

Current efforts such as ReACT \cite{ReACT-iclr23} and PlaSma \cite{PlaSma-iclr24} that rely on explicit plans face a key limitation: specific and question-oriented instructions often fail to generalize well across diverse problems, and the detailed content involved is prone to errors, making it difficult for the LLM to learn effectively. Here, we introduce the idea of mapping plans from language space to latent space to capture high-level, generalizable, and concise instructions that provide conceptual-level reasoning guidance. Unlike explicit plans tailored to individual problems, their encodings are not question-specific; instead, they offer general guidance applicable across a variety of contexts. As a result, despite differing textual formulations, the encodings retain only the commonality that generalizes well across problems.

To show these benefits, we prompt DeepSeek-V3 \cite{DeepSeek-V3} to perform \(\vc_{1\dots n} \sim f_{\bm{\theta}} \left(\vc_{1\dots n} \mid Q, \bm{A}, \rmD\right)\) and \(\bm{p}_{1\dots n} \sim f_{\bm{\theta}} \left(\bm{p}_{1\dots n} \mid Q, \vc_{1\dots n}, \rmS\right)\), where $\bm{A}$ denotes the CoT answer, while $\rmD$ and $\rmS$ represent the prompts for answer decomposition and explicit plan summarization. We present the results in Figure~\ref{fig:comparsion}. Specifically, after removing stop words from the questions and the summarized skeleton plans using NLTK, we encode them with the \textit{all-MiniLM-L6-v2} model. We then visualize the pairwise distances with a heatmap and project the embeddings into 2D using t-SNE.

Figure~\ref{fig:comparsion} shows that embeddings of explicit plans exhibit a certain level of commonality and are capable of generalizing across problems. Using Figure~\ref{fig:question}, which visualizes question similarity across seven categories, as a reference, we observe that explicit plans from reasoning Step 1 and Step 3 (Figure~\ref{fig:explicitplan1}) reveal two key trends: (1) when two questions are similar, their corresponding plans also tend to be close in embedding space; and (2) as the reasoning step increases, the degree of commonality strengthens. These trends are further supported by the clustering patterns of plans from Step 1 and Step 3 shown in Figures~\ref{fig:explicitplan1cluster} and~\ref{fig:explicitplan3cluster}: plans across different categories not only exhibit clear boundary separation but also significant overlap, indicating that a single plan can be applicable to questions both within and across categories.

\section{Methodology}
\label{sec:method}

\begin{figure*}[t]
    \centering
    
    \begin{subfigure}[b]{\textwidth}
        \includegraphics[width=\linewidth]{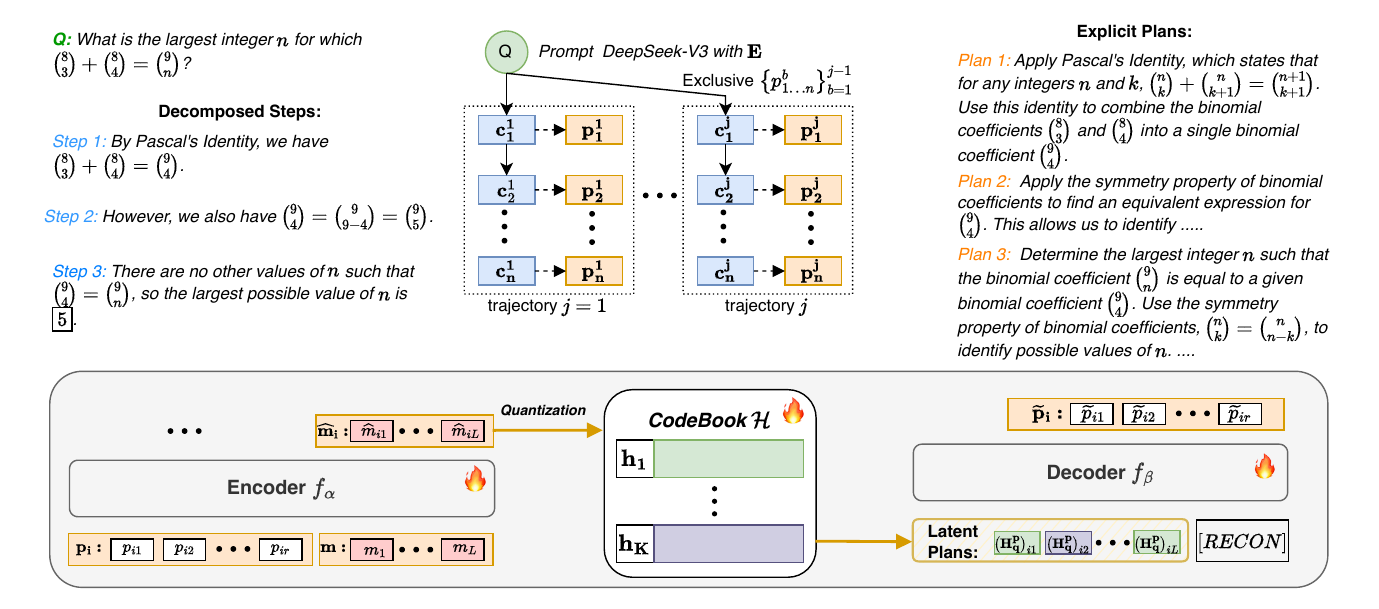} 
    \end{subfigure}
    \caption{Illustration of the overall pipeline of \emph{iCLP}. The upper part shows the process of Plan Distillation using a sample from the Counting \& Probability category of the \texttt{MATH} dataset. The right part depicts the encoder - quantizer - decoder structure used for Latent Plan Generation.}
    \label{fig:main-structure}
\end{figure*}

This section introduces the three components of our framework, \emph{iCLP}: distilling explicit plans from existing answers, learning a latent plan space, and finetuning LLMs with latent plans (LPs). The overall pipeline is illustrated in Figure~\ref{fig:main-structure}. Our core objective, motivated by Subsection~\ref{subsec:motivation}, is to capture the commonalities and generalizable reasoning guidance beyond plans to support accurate reasoning across diverse tasks.

\subsection{Explicit Plan Distillation}
\label{subsec:explicitplan}

\emph{iCLP} distills explicit plans, denoted as \(\vp_{1\dots n}\), from existing CoT answers by prompting an off-the-shelf LLM. Specifically, for each question and its corresponding generated answer, as described in Subsection~\ref{subsec:motivation}, \emph{iCLP} first decomposes the answer into $n$ separate reasoning steps, marking the \(i\)-th step with the phrase ``Step \(i\)''. It then summarizes a plan for each step using zero temperature decoding, resulting in $n$ plans \(\vp_{1\dots n}\), one for each step.

To further increase the diversity of plans, in addition to using the provided answer for each sample, we prompt the LLM to generate \(U\) CoT reasoning trajectories per question. To prevent the LLM from reusing existing plans, for any reasoning trajectory indexed by \(j \in \left[2,\dots,U\right]\), we use a prompting formulation given by \(\vc_{1\dots n} \sim f_{\bm{\theta}} \left(\vc_{1\dots n} \mid Q, \rmE, \left\{\vp^b_{1\dots n}\right\}^{j-1}_{b=1}\right)\). Here, \(\rmE\) is a textual prompt that instructs the LLM not to reuse any previously used chains of plans \(\left\{\vp^b_{1\dots n}\right\}^{j-1}_{b=1}\) during reasoning. Importantly, \(\rmE\) permits the repetition of individual plans but not of entire chains \(\vp^b_{1\dots n}\) because one chain can encode specific reasoning logic. After filtering out CoT trajectories that lead to incorrect solutions, we retain \(\vp^{U^{\prime}}_{1\dots n}\) for each question. Note that while the value of \(n\) varies across questions, we use fixed notation here for simplicity.

\subsection{Latent Plan Generation}

Although explicit plans present clear instructions and generalize among similar questions, they remain susceptible to token-level errors, particularly when LLMs hallucinate. To address this, we propose learning a latent plan space, which serves as a hidden representation of the plans. By conducting planning in this latent space, we mitigate the negative impact of token-level errors and further enhance the LLM's generalization ability, as the reasoning no longer depends on explicit and specific textual instructions.

To construct this latent space, we first follow the ICAE \cite{icae-iclr24} to introduce few memory tokens, denoted as \(\vm = \left[m_1,\dots,m_L\right]\), where \(L\) is the length. Then, we introduce a vector-quantized autoencoder comprising an encoder \(f_{\bm{\alpha}}\), a discrete codebook \(\gH = \left\{\vh\right\}_{k=1}^K \subset \sR^{d_h}\) and a decoder \(f_{\bm{\beta}}\), where \(d_h\) represents the dimensionality of the code. Similar to the pipeline in VQ-VAE \cite{vqvae-neurips17} and VQ-GAN \cite{vqgan-cvpr21}, we approximate a given plan \(\vp\) by \(\widetilde{\vp}=f_{\bm{\beta}}\left(\mH^{\vp}_{\bm{q}}\right)\), where \(\mH^{\vp}_{\bm{q}} \in \sR^{d_h \times |\vp|}\) is a spatial collection of codebook entries representing \(\vp\). Specifically, we set the \(f_{\bm{\alpha}}\) to be an encoder-only transformer and \(f_{\bm{\beta}}\) to be a decoder-only transformer. Thus, we obtain the \(\mH^{\vp}_{\bm{q}}\) by performing element-wise quantization \(\bm{q}\left(\widehat{\mH}^{\vp}\right)\), where each spatial code \(\widehat{\bm{h}}_{L} \in  \sR^{d_h}\) from only the encoded memory representation \(\widehat{\mH}^{\vp} = f_{\bm{\alpha}}\left(\left[\vp, \vm\right]\right) \in \sR^{d_h \times |\vm|}\) is mapped to its nearest codebook entry \(\bm{h}_{k}\), where \(e \in [1, \dots, |\vm|]\) here is the token index of the memory token. This process is formulated as follows:

\begin{equation}
\begin{split}
     &\mH^{\vp}_{\bm{q}} = \bm{q}\left(\widehat{\mH}^{\vp}\right) := \argmin_{\bm{h}_{k} \in \gH} ||\bm{h}_{k}-\widehat{\bm{h}}_{e}||, \\ 
     &\widetilde{\vp}=f_{\bm{\beta}}\left(\bm{q}\left(f_{\bm{\alpha}}\left(\left[\vp, \vm\right]\right) \right)\right), \forall e \in [1, \dots, |\vm|]. 
\end{split}
\end{equation}

For the reconstruction task, where \(\widetilde{\vp} \approx \vp\), we employ the completion loss, as that presented in ICAE \cite{icae-iclr24}. Given the quantized representation \(\mH^{\vp}_{\bm{q}}\) and a special reconstruction indicator token `\(\left[RECON\right]\)', the decoder predicts the input \(\vp\) via next-token prediction. To optimize \(\bm{\alpha}, \gH, \bm{\beta}\), we propagate gradients from the decoder to the encoder, avoiding the non-differentiable quantization. The combination of the cross-entropy loss and commitment loss is computed as: \(-\frac{1}{T} \sum_{t=1}^{T} \log P_\theta(x_t \mid x_{<t}) + ||sg\left[f_{\bm{\alpha}}\left(\vp\right)\right]-\mH^{\vp}_{\bm{q}}||^2+||sg\left[\mH^{\vp}_{\bm{q}}\right]-f_{\bm{\alpha}}\left(\vp\right)||\). With the trained \(\bm{\alpha}, \gH\), we can obtain the \textbf{latent plan} for any plan by representing it with \(\mH^{\vp}_{\bm{q}}\).

\subsection{Fine-tuning LLMs for Latent Planning}
\label{subsec:finetune}

We synthesize new samples to fine-tune the LLM for planning in the latent space to enhance reasoning in the language space. Specifically, for a sample \(q, \vc_{1 \dots n}\) with distilled plans \(\vp_{1 \dots n}\), we first compute each plan encoding as \(f_{\bm{\alpha}}\left(\left[\vp_{i}, \vm\right]\right)\), then map each memory token encoding to its closest codebook entry in \(\gH\). Thus, we have the matching indexes denoted as \(\gI_{\vp}=\left\{\gI_{\vp_i}\right\}_{i=1}^n\), where for each \(\vp_i\), \({\gI_{\vm_{ie}}} = k\) corresponds to \(\left(\mH^{\vp}_{\bm{q}}\right)_{ie} = \vh_{k}\).

For any LLM, we extend the vocabulary size to include the size of the codebook \(K\) by adding special tokens denoted as `\(\left[LP\{idx\}\right]\)' where \{idx\} corresponds to the row index in the codebook. Consequently, we reformulate the sample into the form ($user$: question, $assistant$: latent plans and CoTs). This is obtained by first expressing the sample as ($user$: question, $assistant$: \(\vp_1, \vc_1, \dots, \vp_n, \vc_n\)) following by replacing the token IDs of each \(\vp_i\) with \(\gI_{\vm_{i}}\). As a result, with supervised fine-tuning, we train the extended vocabulary embeddings and fine-tune the LLM using the completion loss.

\section{Experiments}
\label{sec:exps}

\textbf{Datasets}. For the mathematical task, we use the \texttt{MATH} \cite{MATH-arxiv21} and \texttt{GSM8K} \cite{GSM8K-arxiv21} datasets for model fine-tuning, while the \texttt{AIME2024} \cite{MAA:AIME} and \texttt{MATH-500} \cite{MATH-arxiv21} datasets are used exclusively for evaluation. For the code generation task, we use the \texttt{CodeContests} \cite{CodeContests} (3760 training, 165 test) dataset for fine-tuning, and the \texttt{HumanEval} \cite{HumanEval} and \texttt{MBPP} \cite{MBPP} datasets for evaluation.  

\textbf{Learning Settings}. We use Qwen2.5 \cite{Qwen2.5} in different sizes including 0.5B, 3B, 7B as the base models for fine-tuning toward latent planning. For all cases, to generate new CoT answers or extract the explicit plans, we prompt DeepSeek-V3 with a zero temperature and 0.3 temperature, respectively. For supervised fine-tuning, the learning rates for the 0.5B, 0.6B, 1.7B, 3B, and 7B models are set to 2e-5, 2e-5, 1e-5, 8e-6, and 5e-6, respectively. We employ the AdamW optimizer with a cosine learning rate scheduler. For \emph{iCLP}, the encoder is the all-MiniLM-L6-v1 model from sentence-transformers, while the decoder is Qwen2.5-3B with the extended vocabulary, as discussed in subsection \ref{subsec:finetune}. For the quantizer, the size of the codebook is 2048, the dimension is 512, and $\beta$ is 0.3. The training of \emph{iCLP} uses LoRA \cite{LoRA-iclr21}, with a batch size of 16 for 2 epochs. For the Plan Distillation, we set the \(U\) to be 20. Throughout the experiment, we set the number of memory tokens to \(L=6\) due to that the explicit plan is generally a short sentence.

\textbf{Baselines}. In addition to comparisons with base LLMs, \emph{iCLP} is evaluated against state-of-the-art (SOTA) fine-tuning (FT) methods, including Learn Planning and Reasoning \cite{LearnPlanningReasoning-emnlp24}, PS+/PS \cite{ps-acl23}, ReAct \cite{ReACT-iclr23}, PlaSma \cite{PlaSma-iclr24}, and Coconut \cite{Coconut-arxiv24}. However, fully reproducing these methods is relatively infeasible due to their strong dependence on specific task settings. Instead, we implement their core ideas within the context of our work, which involves step-wise explicit plans. These approaches can be viewed as instances of fine-tuning (FT) large language models on plan-based reasoning samples synthesized from existing datasets. We refer to this setting as FT-explicit (FT-E), where each step in the synthesized data includes a corresponding explicit plan. We also introduce Coconut, which implements latent chain-of-thought reasoning. This approach is distinct compared to our latent planning method and is therefore referred to as FT-special (FT-S). Additionally, for ablation analysis, we apply \emph{iCLP} to learn a latent representation of the explicit plans and fine-tune the model based on this latent space, denoted as the \textit{FT-latent explicit} (\textit{FT-LE}) setting. \emph{iCLP} is the full version of our method, which fine-tunes LLMs using latent plans learned from explicit plans. In particular, we compare our method with the state-of-the-art (SOTA) approach GRPO \cite{GRPO}, which fine-tunes LLMs using reinforcement learning.

\textbf{Metrics}. We report the pass\@1 accuracy (in \%) of the LLM evaluated in three modes: In the \textit{normal mode}, the LLM with \emph{iCLP} is evaluated directly on the test set from the same dataset as the train set. In the \textit{cross mode}, the LLM with \emph{iCLP} is evaluated on a test set from a dataset different from the one used for training, denoted as training dataset $\rightarrow$ test set. The \textit{accumulation mode} involves collecting explicit plans from multiple datasets to train the LLM with \emph{iCLP}, followed by evaluation on various datasets.

\subsection{Main Results}
\label{subsec:mainresults}

\begin{table*}[tbp]
    \footnotesize
    \centering
    \caption{Evaluating the reasoning performance of a series of Qwen2.5 models with different methods. The abbreviates of the datasets \texttt{MATH}, \texttt{MATH-500}, \texttt{AIME2024}, \texttt{GSM8K}, \texttt{CodeContests}, and \texttt{HumanEval} are \texttt{M}, \texttt{M-500}, \texttt{AM}, \texttt{G8}, \texttt{CC}, \texttt{HE}, and \texttt{MBPP}, respectively. The $\times$ indicates that, under the corresponding fine-tuning method for LLMs, training fails to converge. The $\rightarrow$ indicates cross-dataset evaluation, as described in the `Metrics' part of the experimental settings. Here, the base refers to LLMs using zero-shot CoT prompting.}
    \begin{adjustbox}{max width=\textwidth}
    \begin{tabular}{c|c|ccccccccc}
    \toprule
    \multirow{2}{*}{Qwen2.5} & \multirow{2}{*}{Methods} & \multicolumn{3}{c|}{\textit{Normal Mode}}                                  & \multicolumn{6}{c}{\textit{Cross Mode}}                                                                                                                                                                                                                                 \\ 
                            &                           & \texttt{M}          & \texttt{G8}         & \multicolumn{1}{c|}{\texttt{CC}} & \multicolumn{1}{c|}{\texttt{M}$\rightarrow$\texttt{AM}} & \multicolumn{1}{c|}{\texttt{G8}$\rightarrow$\texttt{AM}} &  \multicolumn{1}{c|}{\texttt{M}$\rightarrow$M-500} & \multicolumn{1}{c|}{\texttt{G8}$\rightarrow$\texttt{M-500}}  & \texttt{CC}$\rightarrow$\texttt{HE} & \multicolumn{1}{c}{\texttt{CC}$\rightarrow$\texttt{MBPP}} \\ \hline\hline
                            \multirow{5}{*}{0.5B}   
                            & Base                      & 19.5          & 41.6          & 1.2                               & 0                           & 0                                                 & 15.8                        & 15.8                                              & 30.5                                 & 39.3                                                 \\ 
                            & GRPO                      & \textbf{49.6}          & \textbf{54.5}          & \(\times\)                               & 0                           & 0                                                 & \textbf{34.4}                        & 17.9                                              & \(\times\)                                &  \(\times\)                                                 \\ \cline{2-2}
                            & \textit{FT-E}               & 23.1          & 43.8          & 5.5                               & 0                           & 0                                                 & 18.8                        & 15.8                                              & 32.9                                 & 43.6                                                 \\ 
                            & \textit{FT-S}               & 31.1          & 48            & 8.5                               & 0                           & 0                                                 & 27.2                        & 18.4                                              & 36.6                                 & 47                                                   \\ 
                            & \textit{FT-LE}           & 13.3          & 38.3          & 1.8                               & 0                           & 0                                                 & 11.2                        & 9.6                                               & 19                                   & 21.4                                                 \\ 
                            & \bf{iCLP}                      & 36.7 & 51.5 & \textbf{11.5}                              & 0                  & 0                                        & 28.1               & \textbf{20.8}                                     & \textbf{39.6}                        & \textbf{51}                                          \\ \hline
    \multirow{5}{*}{3B}     
                            & Base                      & 42.6          & 79.1          & 9                                 & 0                           & 0                                                 & 39.6                        & 39.6                                              & 42.1                                 & 57.1                                                 \\ 
                            & GRPO                      & \textbf{68.5}          & \textbf{86.6}          & \(\times\)                               & 10                           & 0                                                 & \textbf{60.8}                        & 40.4                                              & \(\times\)                                &  \(\times\)                                                 \\ \cline{2-2}
                            & \textit{FT-E}               & 48.2          & 81.4          & 10.9                              & 0                           & 0                                                 & 43                          & 39.8                                              & 44.5                                 & 61.2                                                 \\ 
                            & \textit{FT-S}               & 54.9          & 83.2          & 13.9                              & 7.6                         & 0                                                 & 48.4                        & 40                                                & 47                                   & 65.2                                                 \\ 
                            & \textit{FT-LE}           & \(\times\)             & \(\times\)             & 12.1                              & \(\times\)                           & 0                                                 & \(\times\)                           & \(\times\)                                                 & 44.5                                 & 62                                                   \\ 
                            & \bf{iCLP}                      & 60.1 & 85   & \textbf{18.8}                              & \textbf{10}                 & 0                                        & 55.4               & \textbf{44.2}                                     & \textbf{54}                          & \textbf{73}                                          \\ \hline
    \multirow{5}{*}{7B}     
                            & Base                      & 49.8          & 85.4          & 15.8                              & 3.3                         & 3.3                                               & 42.4                        & 42.4                                              & 53                                   & 74.9                                                 \\ 
                            & GRPO                      & \textbf{83.7}          & \textbf{91.7}          & 19.2                               & 20                           & 3.3                                                 & \textbf{78.2}                        & 45.6                                              & 53.2                                &  76                                                 \\ \cline{2-2}
                            & \textit{FT-E}               & 57.4          & 87.3          & 19.4                              & 10                          & 3.3                                               & 52                          & 43.8                                              & 56.5                                 & 75.6                                                 \\ 
                            & \textit{FT-S}               & 65.8          & 88.9          & 22.4                              & 16.7                        & 3.3                                               & 58.4                        & 44                                                & 59.5                                 & 78.9                                                 \\ 
                            & \textit{FT-LE}           & \(\times\)             & \(\times\)             & \(\times\)                                 & \(\times\)                           & \(\times\)                                                 & \(\times\)                           & \(\times\)                                                 & \(\times\)                                    & \(\times\)                                                    \\ 
                            & \bf{iCLP}                      & 74.3 & 90.1 & \textbf{26.7}                              & \textbf{20}                 & \textbf{3.3}                                      & 68.6               & \textbf{46.2}                                     & \textbf{66.5}                        & \textbf{86.9}                                        \\     
    \bottomrule
    \end{tabular}
    \end{adjustbox}
    \vspace{-4mm}
    \label{tab:mainresults}
\end{table*}

Table~\ref{tab:mainresults} shows that LLMs with \emph{iCLP} significantly enhance reasoning performance across three datasets, demonstrating the ability to perform \textbf{generalizable planning and achieve high accuracy} across diverse tasks. In the \textit{normal mode} and the more challenging \textit{cross mode}, \emph{iCLP} consistently outperforms all baselines and closely matches the performance of the GRPO algorithm \cite{GRPO}. These results suggest that reasoning with latent planning enables LLMs to acquire reasoning abilities that are both reliable and generalize effectively across a wide range of problem-solving tasks.

Specifically, \emph{iCLP} consistently outperforms Base in \textit{normal mode}, with an average accuracy improvement of 11.5. Although GRPO shows better performance on \texttt{MATH} and \texttt{GSM8K}, with average gains of 8 at 0.3B, 5 at 3B, and 5.5 at 7B, \emph{iCLP} achieves the strongest results on \texttt{CodeContests}, where GRPO fails to converge at 0.5B and 3B and falls behind by 7.5 at 7B. For math reasoning tasks on \texttt{MATH} and \texttt{GSM8K}, \emph{iCLP} surpasses \textit{FT-E}, \textit{FT-S}, and \textit{FT-LE} with average improvements of 9.4, 4.3, and 30.4 respectively. On \texttt{CodeContests}, the gains are 7.1 over \textit{FT-E}, 4.1 over \textit{FT-S}, and 18.4 over \textit{FT-LE}. These results suggest that \emph{iCLP} enables LLMs of various sizes to perform latent-space planning as generalized guidance, resulting in substantial improvements in both mathematical reasoning and code generation. 

\begin{table}[tbp]
\centering
\caption{Evaluation of the average generation token cost across different methods on \texttt{MATH} and \texttt{TheoremQA}. We report both the average and standard deviation (mean $\pm$ std) of the total tokens used per question, including tokens used for prompting the LLMs and those generated by the models.}
\label{tab:token_cost}
\begin{adjustbox}{width=\columnwidth}
\begin{tabular}{c|c|c}
\toprule
\textbf{Methods} & \texttt{MATH} & \texttt{TheoremQA} \\
\hline\hline
0.5B w/ ZeroCoT & 221.6 $\pm$ 172.5 & 250.3 $\pm$ 110.2 \\
14B w/ ZeroCoT & 261.8 $\pm$ 192.2 & 308.7 $\pm$ 137.5 \\
0.5B w/ PS+ & 327.5 $\pm$ 176.7 & 367.5 $\pm$ 153.6 \\
0.5B w/ Prompting & 815.2 $\pm$ 356.7 & 993.4 $\pm$ 390.5 \\
0.5B w/ \emph{iCLP} & 250.3 $\pm$ 110.9 & \textbf{200.7 $\pm$ 100.2} \\
7B w/ ZeroCoT & 246.9 $\pm$ 189.5 & 291.2 $\pm$ 160.8 \\
7B w/ PS & 357.2 $\pm$ 200.9 & 390.6 $\pm$ 190 \\
7B w/ Prompting & 934.5 $\pm$ 390.2 & 1103 $\pm$ 487.5 \\
7B w/ \emph{iCLP} & 300.6 $\pm$ 150.8 & \textbf{270.2 $\pm$ 127.1} \\
\bottomrule
\end{tabular}
\end{adjustbox}
\vspace{-4mm}
\end{table}
As shown in Table~\ref{tab:token_cost}, Our \emph{iCLP} achieves high \textbf{token efficiency}, with an average token cost of 250.3 $\pm$ 110.9 on \texttt{MATH}. This cost is significantly lower than that of planning-based reasoning methods such as PS+ \cite{ps-acl23}. . In particular, on the more challenging dataset \texttt{TheoremQA}, the token usage is only 200.7 $\pm$ 100.2 for Qwen2.5-0.5B and 370.2 $\pm$ 127.1 for Qwen2.5-7Bwhich is lower even than ZeroCoT. This efficiency arises from two factors: (1) the latent plan requires only 6 tokens, and (2) the plan provides clear guidance, enabling the LLM to avoid unnecessary exploration during reasoning. 

For the \textit{cross mode} for generalizable planning evaluation, we evaluate Qwen2.5 models trained on latent plans derived from one dataset and tested on a different dataset. Qwen2.5 3B and 7B models with \emph{iCLP} trained on \texttt{MATH} successfully solve challenging AIME2024 problems, with the 7B model achieving an accuracy of 20, only 3.3 lower than the much larger Qwen2.5-72B-Instruct. Across all model sizes (0.5B, 3B, 7B), \emph{iCLP} significantly outperforms the Base LLM. While \emph{iCLP} performs slightly below GRPO on \texttt{M}\(\rightarrow\)\texttt{M-500} with an average gap of 3.9, it consistently outperforms GRPO on \texttt{G8}\(\rightarrow\)\texttt{M-500}, with average gains of 2.2, most notably 2.9 at 0.5B and 3.8 at 3B, indicating stronger generalization when transferring plans from \texttt{G8} to \texttt{M-500}. In code generation, \emph{iCLP} shows clear advantages. It improves over Base by an average of 12.4 on \texttt{CC}\(\rightarrow\)\texttt{HE} and \texttt{CC}\(\rightarrow\)\texttt{MBPP}. GRPO fails to converge at 0.5B and 3B on \texttt{CC}, while \emph{iCLP} remains stable and effective. At 7B, where GRPO does converge, \emph{iCLP} still outperforms it by 13.3 on \texttt{CC}\(\rightarrow\)\texttt{HE} and 10.9 on \texttt{CC}\(\rightarrow\)\texttt{MBPP}, demonstrating robust cross-data generalization. Moreover, compared to the variants \textit{FT-E}, \textit{FT-S}, and \textit{FT-LE}, \emph{iCLP} consistently achieves higher accuracy and more stable performance across model sizes and tasks, confirming its effectiveness in cross-data generalization for both math and code reasoning.

\subsection{Continuous Learning with Latent Planning}
\label{subsec:accumulate}

\begin{figure}[t]
    \centering
    
    \begin{subfigure}[b]{0.275\columnwidth}
        \includegraphics[width=\linewidth]{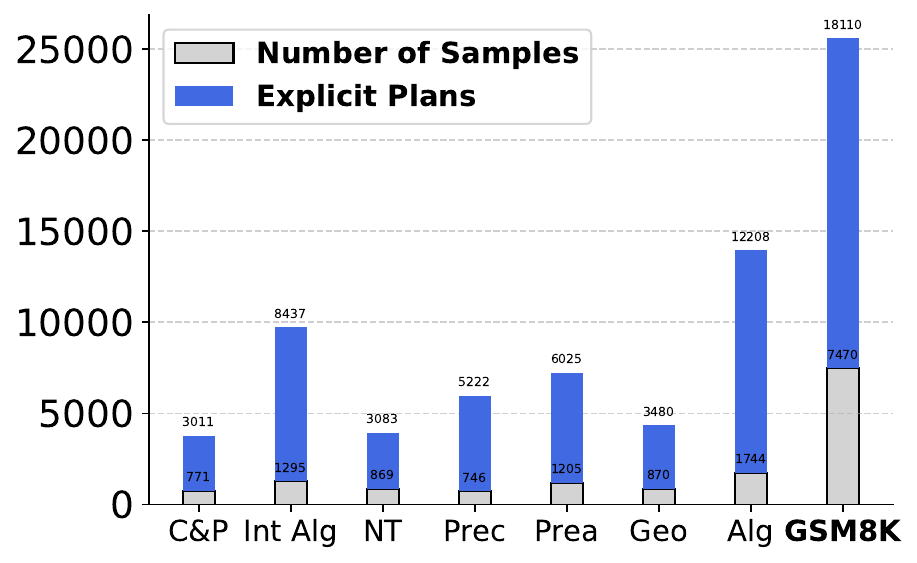} 
        \caption{}
        \label{fig:plancount}
        
    \end{subfigure}
    \hfill
    \begin{subfigure}[b]{0.23\columnwidth}
        \includegraphics[width=\linewidth]{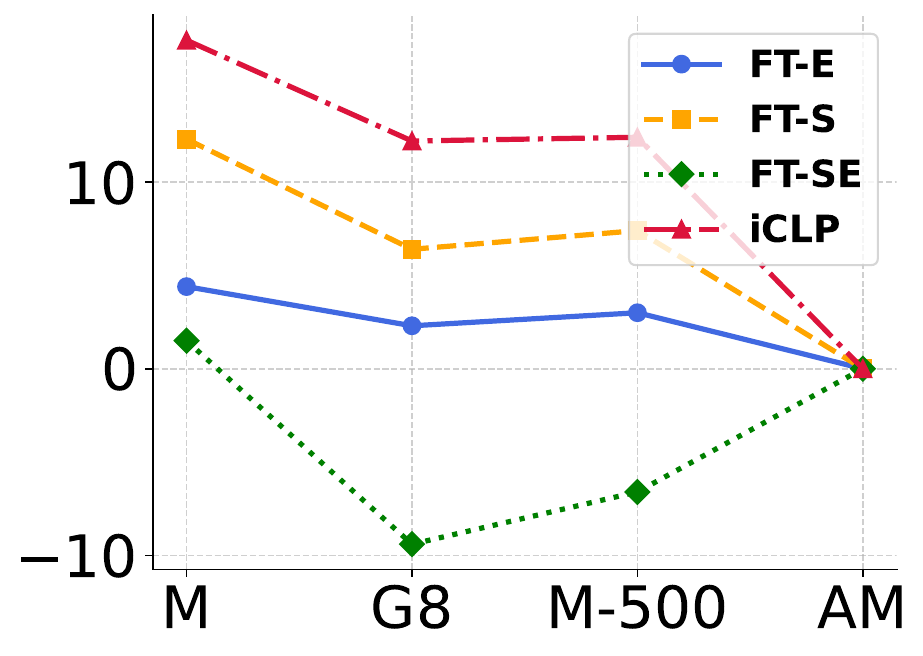} 
        \caption{}
        \label{fig:delta05b}
    \end{subfigure}
    \hfill
    \begin{subfigure}[b]{0.23\columnwidth}
        \includegraphics[width=\linewidth]{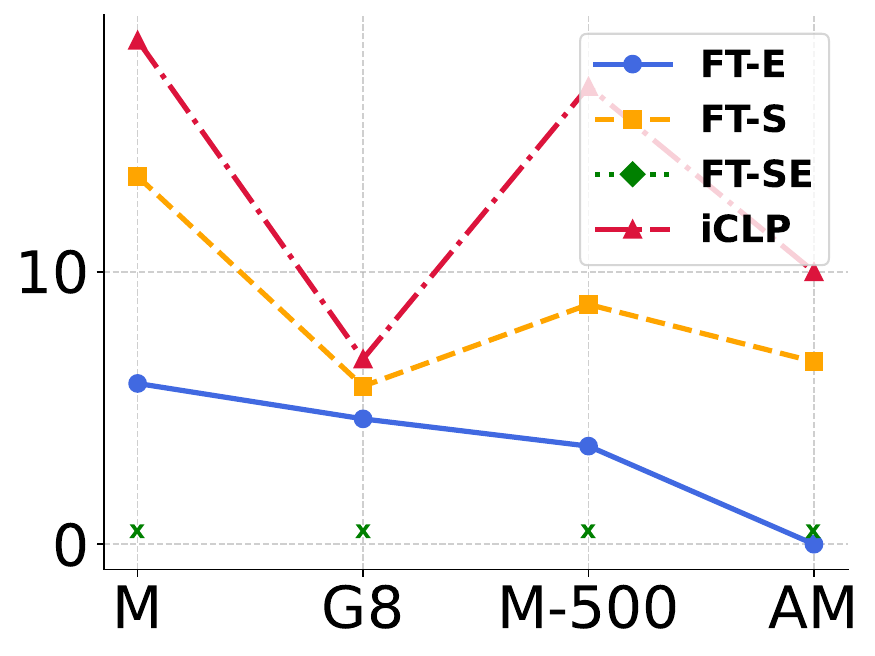} 
        \caption{}
        \label{fig:delta3b}
    \end{subfigure}
    \hfill
    \begin{subfigure}[b]{0.23\columnwidth}
        \includegraphics[width=\linewidth]{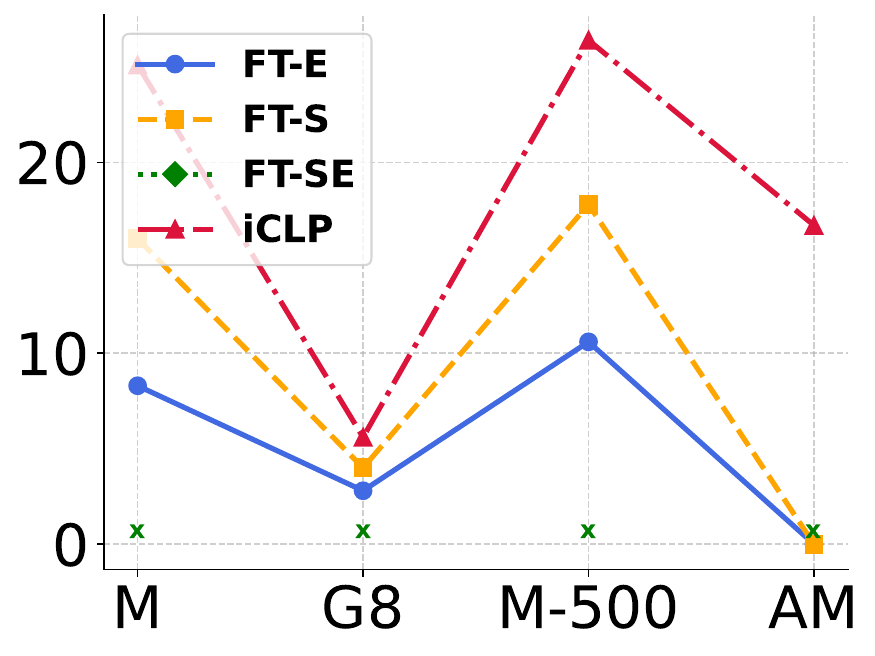} 
        \caption{}
        \label{fig:delta7b}
    \end{subfigure}
    \caption{Illustration of the number of distilled plans and the \textit{accumulation mode} performance of LLMs with \emph{iCLP}. (a) shows the number of explicit plans that can be distilled from each category of the \texttt{MATH} and \texttt{GSM8K} datasets. The x-axis labels represent the abbreviations of the seven category names (see appendix). (b)(c)(d) show the accuracy gain (`y-axis') over the base model after fine-tuning the LLM with plans accumulated from the \texttt{MATH} and \texttt{GSM8K} datasets. Accuracy is measured across four datasets, with abbreviated names provided in Table~\ref{tab:mainresults}. (b)(c)(d) correspond to Qwen2.5 models with 0.5B, 3B, and 7B parameters, respectively.}
    \label{fig:plandelta}
    \vspace{-4mm}
\end{figure}

Figure~\ref{fig:plandelta} demonstrates that in the \textit{accumulation mode}, plans can be progressively accumulated across datasets to enhance the performance of \emph{iCLP} in improving LLM reasoning, thus eavling the continuously learning ability of the LLMs with the latent plan space. Specifically, we first distill plans from the \texttt{MATH} and \texttt{GSM8K} datasets, then merge them and eliminate duplicates based on identical reasoning indices. Using this expanded set of plans, we train the codebook and fine-tune the LLM via \emph{iCLP}. As shown in Figure~\ref{fig:plandelta}, the resulting model is evaluated on four datasets.

Figure~\ref{fig:plancount} highlights the significant difference between the number of explicit plans and the number of questions. In the seven categories of the \texttt{MATH} dataset, the number of explicit plans is not significantly larger than the number of questions, especially in some challenging categories such as \textit{CP}. Besides, LLMs fine-tuned with explicit plans achieve better accuracy, as shown in Table~\ref{tab:mainresults}, indicating that explicit plans effectively capture the commonality and generalizable reasoning patterns across questions. Despite the small scale of explicit plans, such as 18,100 in \texttt{GSM8K}, learning latent plans results in their representation by only a few clusters. As shown in Figure~\ref{fig:iplans2d}, latent plans for questions within the same categories exhibit significant clustering while showing overlap across categories. More importantly, \textbf{fine-tuning LLMs using these latent plans achieves optimal accuracy}, as show in Table~\ref{tab:mainresults}, which approaches SOTA RL-based method GRPO.

Thus, by accumulating plans, we are able to integrate \textbf{generalizable planning knowledge from different datasets}, enabling LLMs trained with \emph{iCLP} to achieve better performance, as shown in Figure~\ref{fig:delta05b}, \ref{fig:delta3b}, and \ref{fig:delta7b}. Across all four evaluation datasets, Qwen2.5 models fine-tuned with latent ones (i.e., \emph{iCLP}) show significant accuracy improvements compared to the base models. These improvements consistently exceed those achieved using explicit plans and \textit{FT-S}. As the Qwen2.5 model size increases from 0.5B to 7B, the performance gains from \emph{iCLP} become more evident, particularly with improvements greater than 20\% on \texttt{MATH} and 10\% on \texttt{AIME2024}.


\subsection{Ablation Study and Qualitative Results}
\label{subsec:qualitative}

Table~\ref{tab:albation} presents the impact of different settings for the encoding dimension and the size of the codebook \(\gH\) on the performance of our \emph{iCLP}. Increasing $d_h$ from 256 to 512 leads to a significant improvement in accuracy, while further increasing it to 1024 offers no additional benefit. Based on this, we set $d_h = 512$ and increase the codebook size $K$ from from 1024 to 2048, which yields accuracy gains of 2.9 and 1.2 points on \texttt{MATH} and \texttt{CodeContests}, respectively. However, further enlarging $K$ does not lead to additional performance improvements. Therefore, based on these results, especially the cross-data evaluation, we argue that the codebook size $K$ plays a more critical role than the dimensionality in improving latent planning performance.

\begin{figure*}[t]
    \centering
    
    \begin{subfigure}[b]{0.24\textwidth}
        \includegraphics[width=\linewidth]{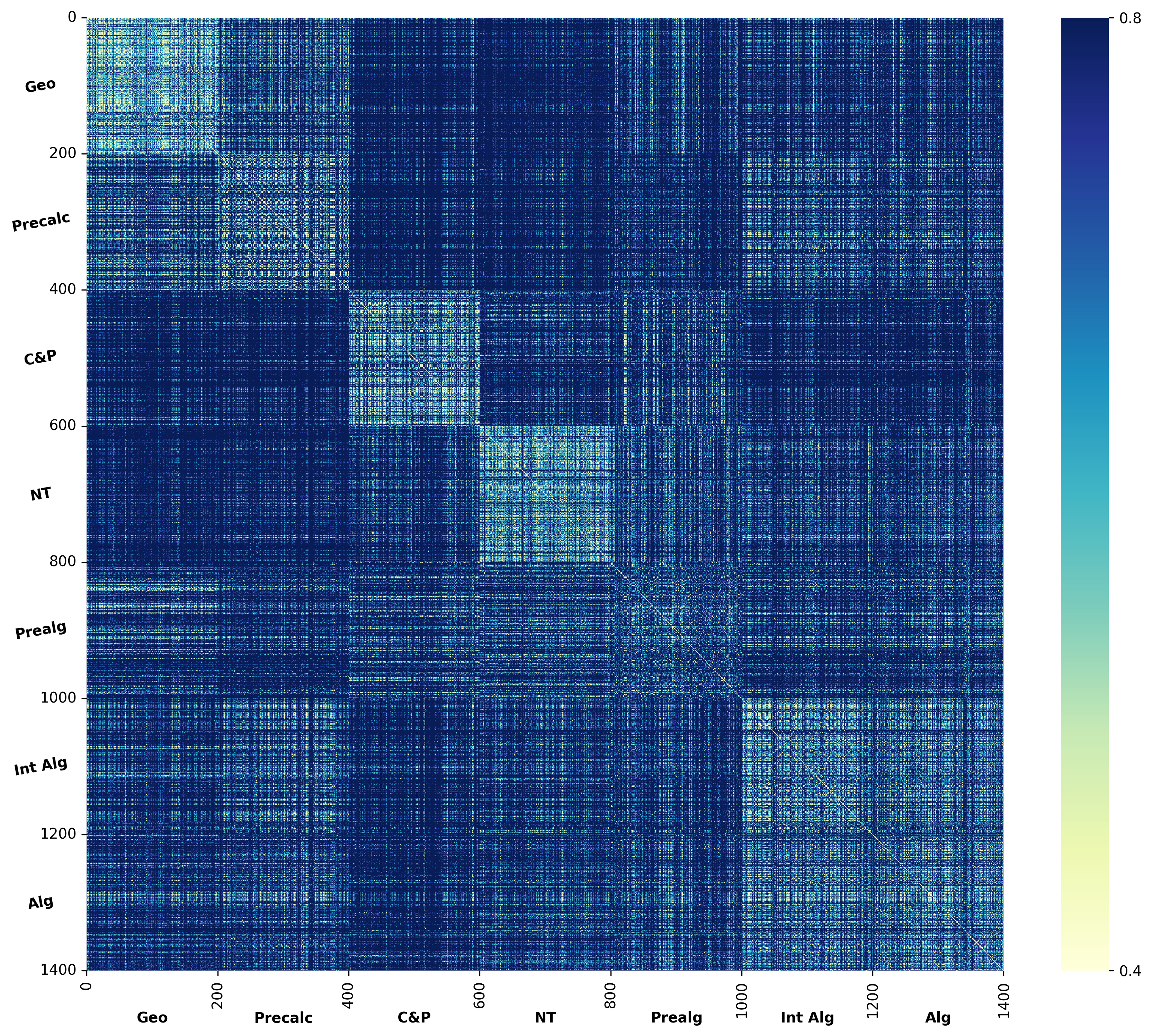} 
        \caption{}
        \label{fig:iplan1relation}
        
    \end{subfigure}
    \hfill
    \begin{subfigure}[b]{0.24\textwidth}
        \includegraphics[width=\linewidth]{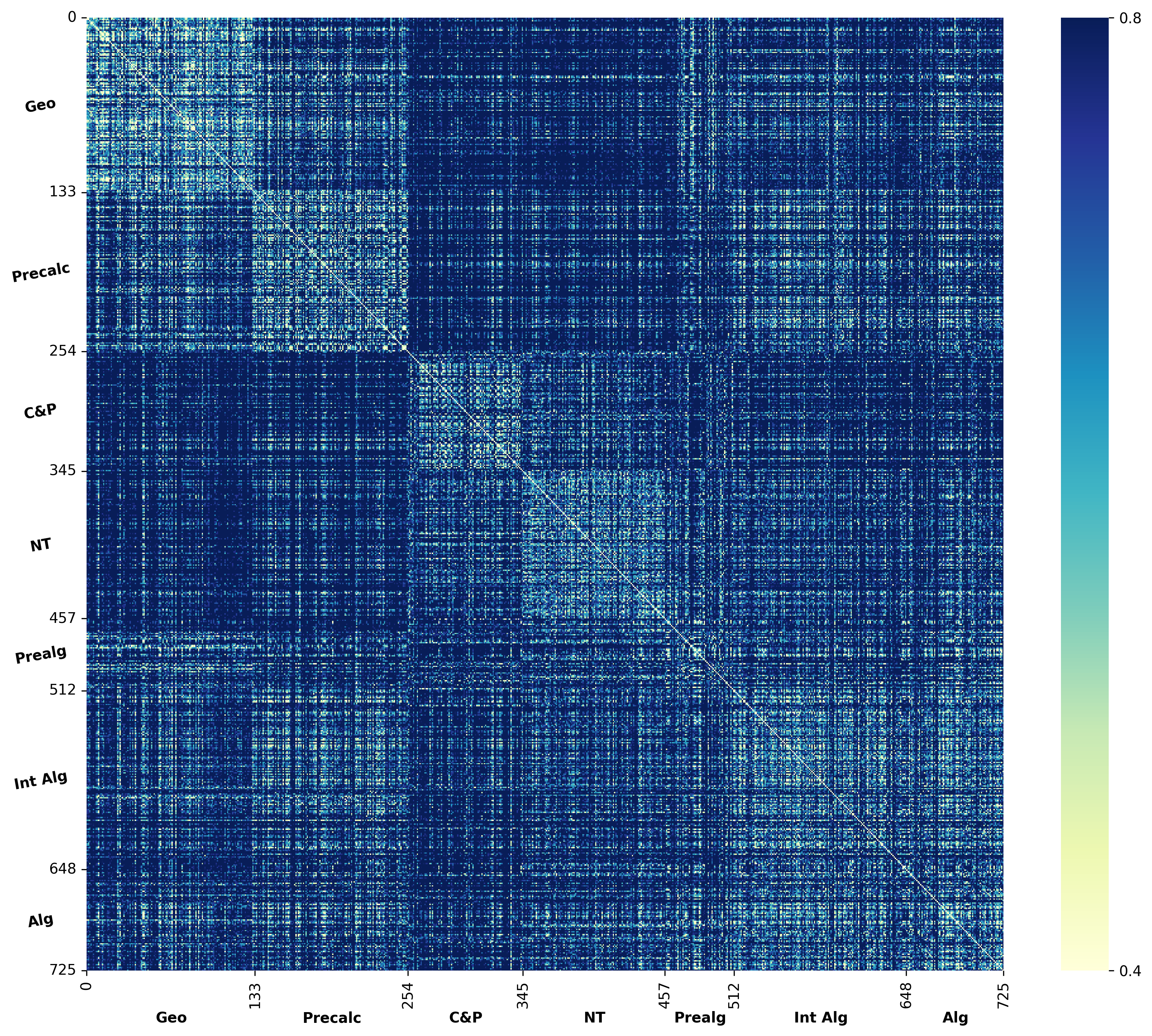} 
        \caption{}
        \label{fig:iplan2relation}
    \end{subfigure}
    \hfill
    \begin{subfigure}[b]{0.24\textwidth}
        \includegraphics[width=\linewidth]{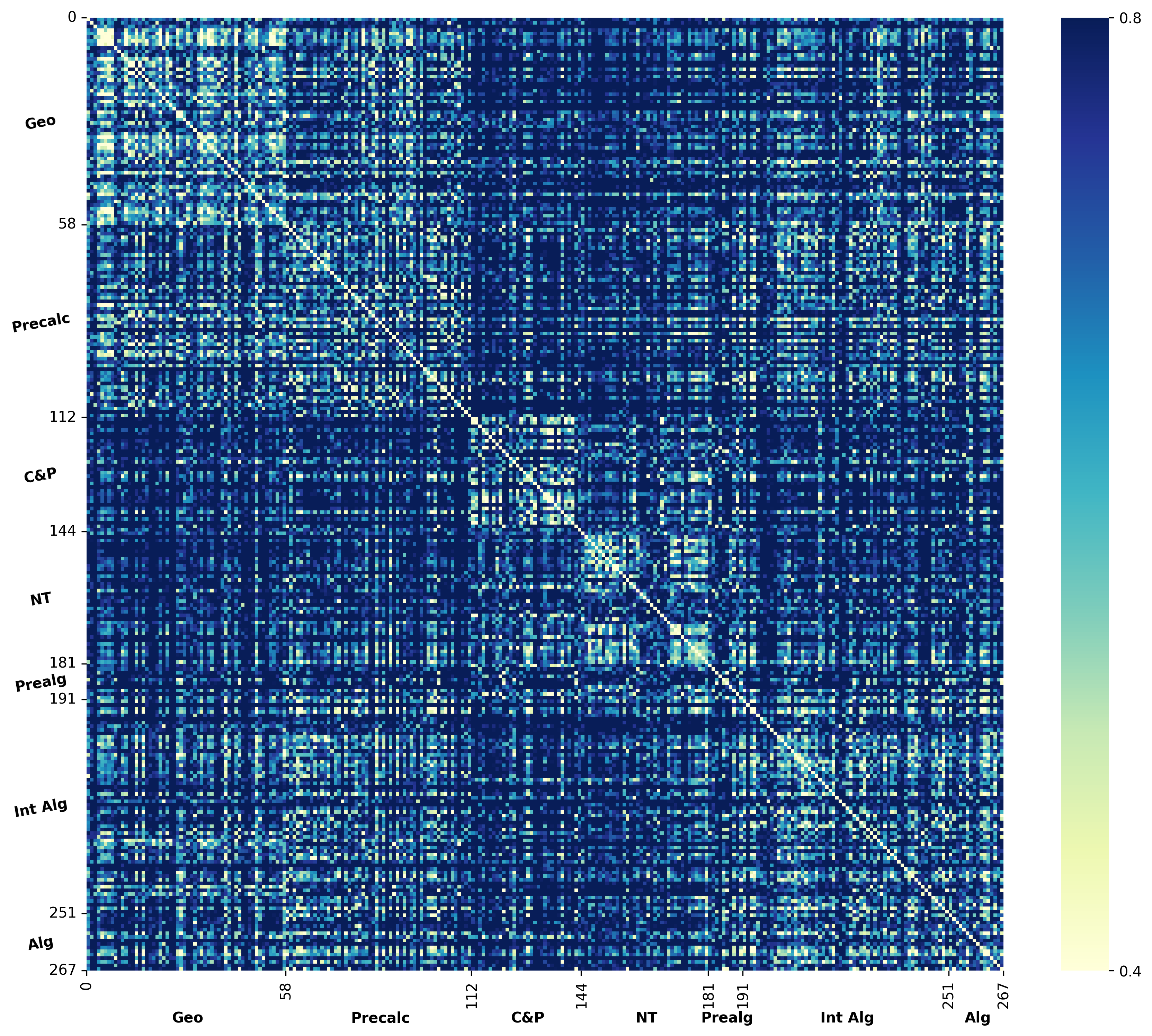} 
        \caption{}
        \label{fig:iplan3relation}
    \end{subfigure}
    \hfill
    \begin{subfigure}[b]{0.24\textwidth}
        \includegraphics[width=\linewidth]{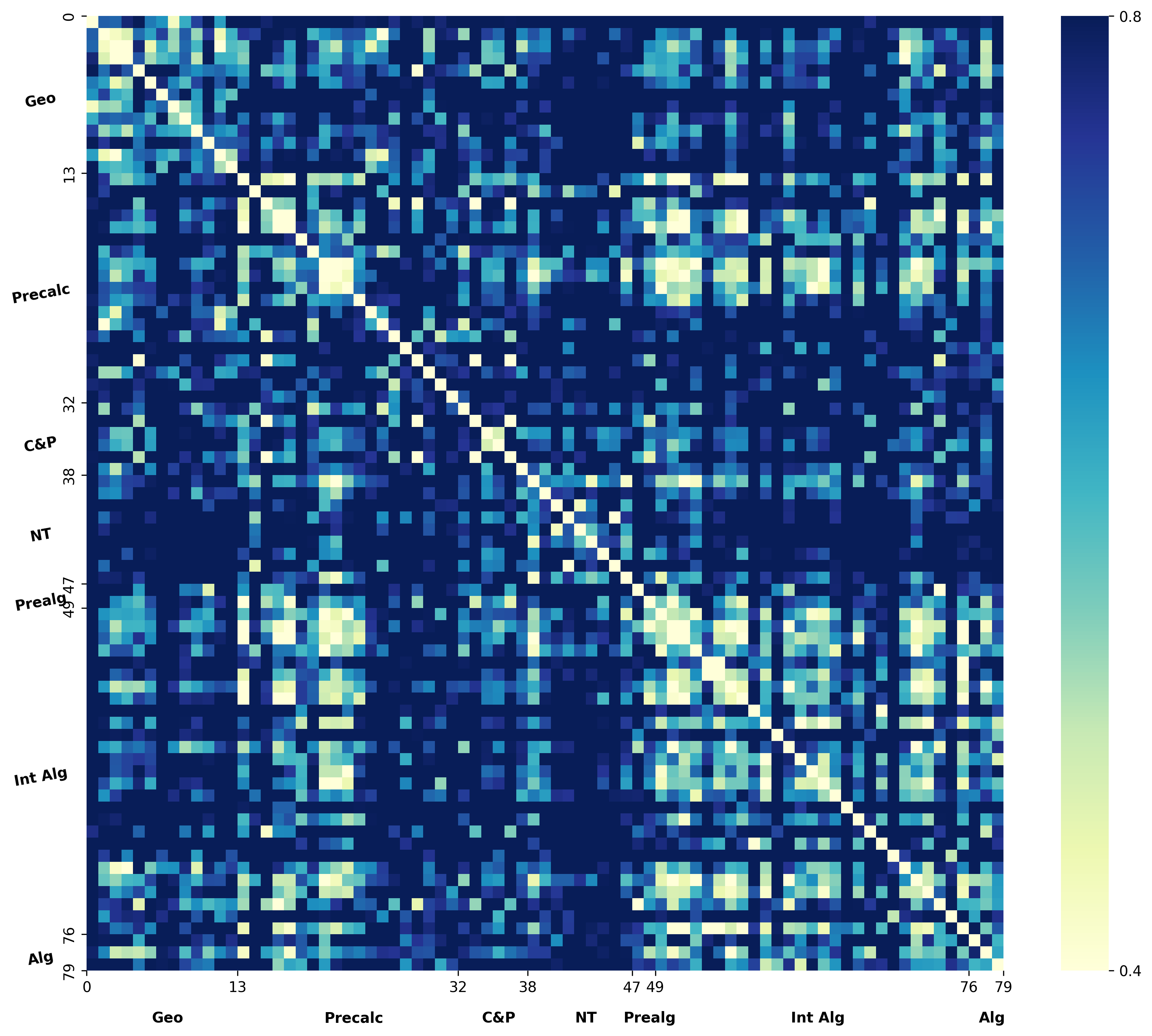} 
        \caption{}
        \label{fig:iplan4relation}
    \end{subfigure}
    \caption{Illustration of the relations of the encoding distances between pairwise latent plans from different reasoning steps (1, 2, 3, and 4). We randomly sample 200 questions from the test set of the \texttt{MATH} dataset and extract the encodings of latent plans generated by Qwen2.5-7B with \emph{iCLP} during problem solving. Subfigures (a), (b), (c), and (d) present the results for the 1st, 2nd, 3rd, and 4th latent plans, respectively.}
    \label{fig:iplanrelations}
    \vspace{-4mm}
\end{figure*}

\begin{figure*}[t]
    \centering

    \begin{subfigure}[b]{0.24\textwidth}
        \includegraphics[width=\linewidth]{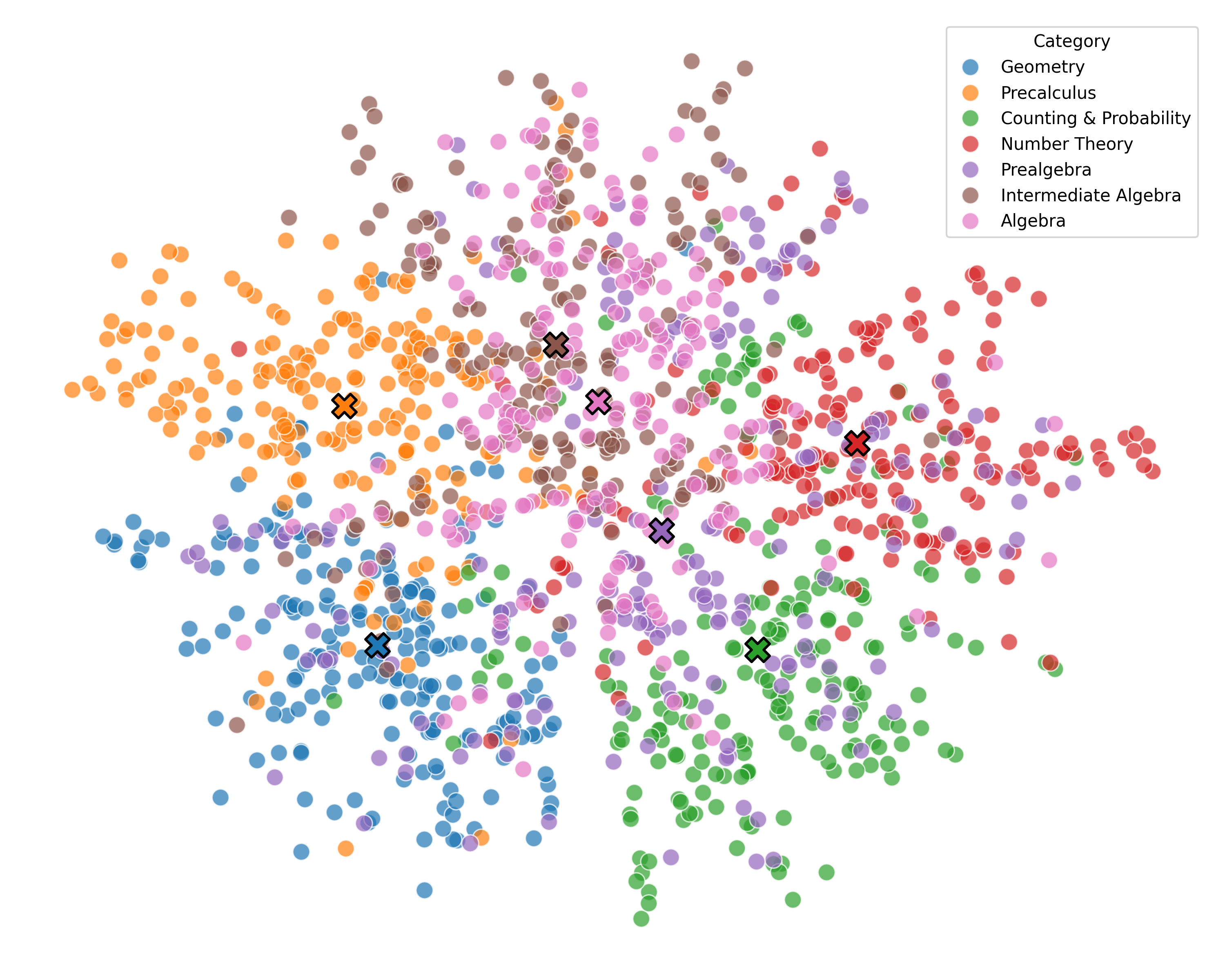} 
        \caption{}
        \label{fig:iplan12d}
        
    \end{subfigure}
    \hfill
    \begin{subfigure}[b]{0.24\textwidth}
        \includegraphics[width=\linewidth]{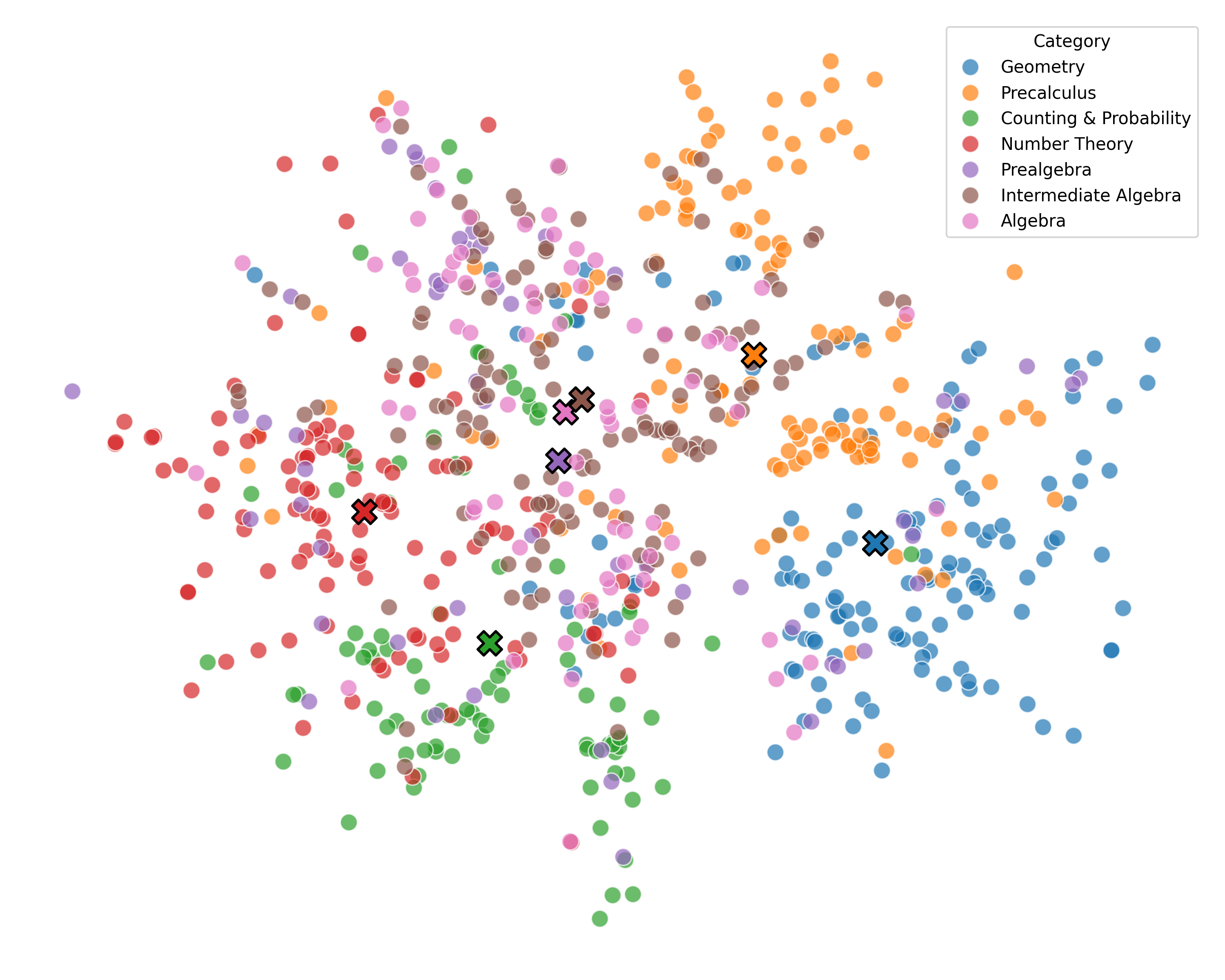} 
        \caption{}
        \label{fig:iplan22d}
    \end{subfigure}
    \hfill
    \begin{subfigure}[b]{0.24\textwidth}
        \includegraphics[width=\linewidth]{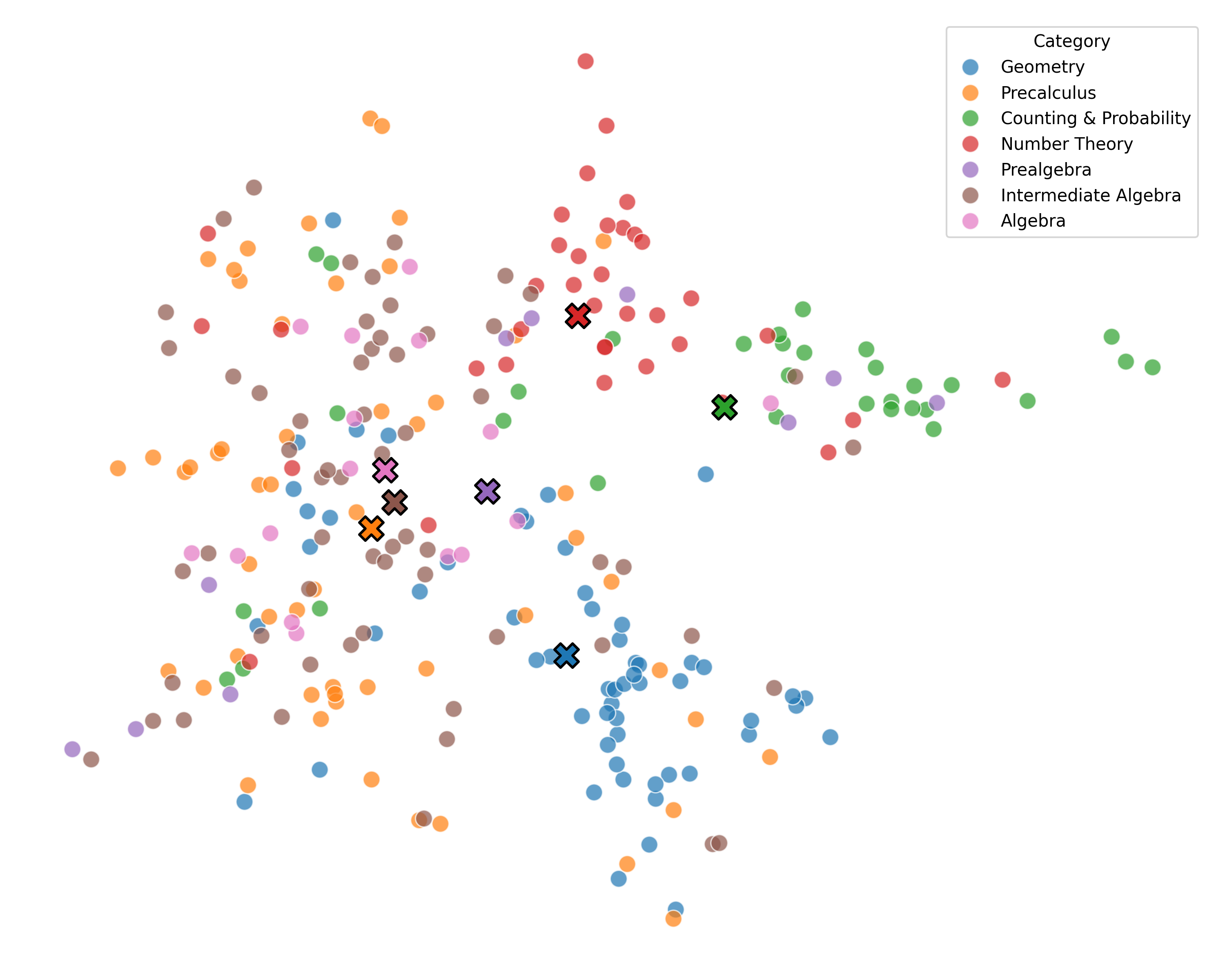} 
        \caption{}
        \label{fig:iplan32d}
    \end{subfigure}
    \hfill
    \begin{subfigure}[b]{0.24\textwidth}
        \includegraphics[width=\linewidth]{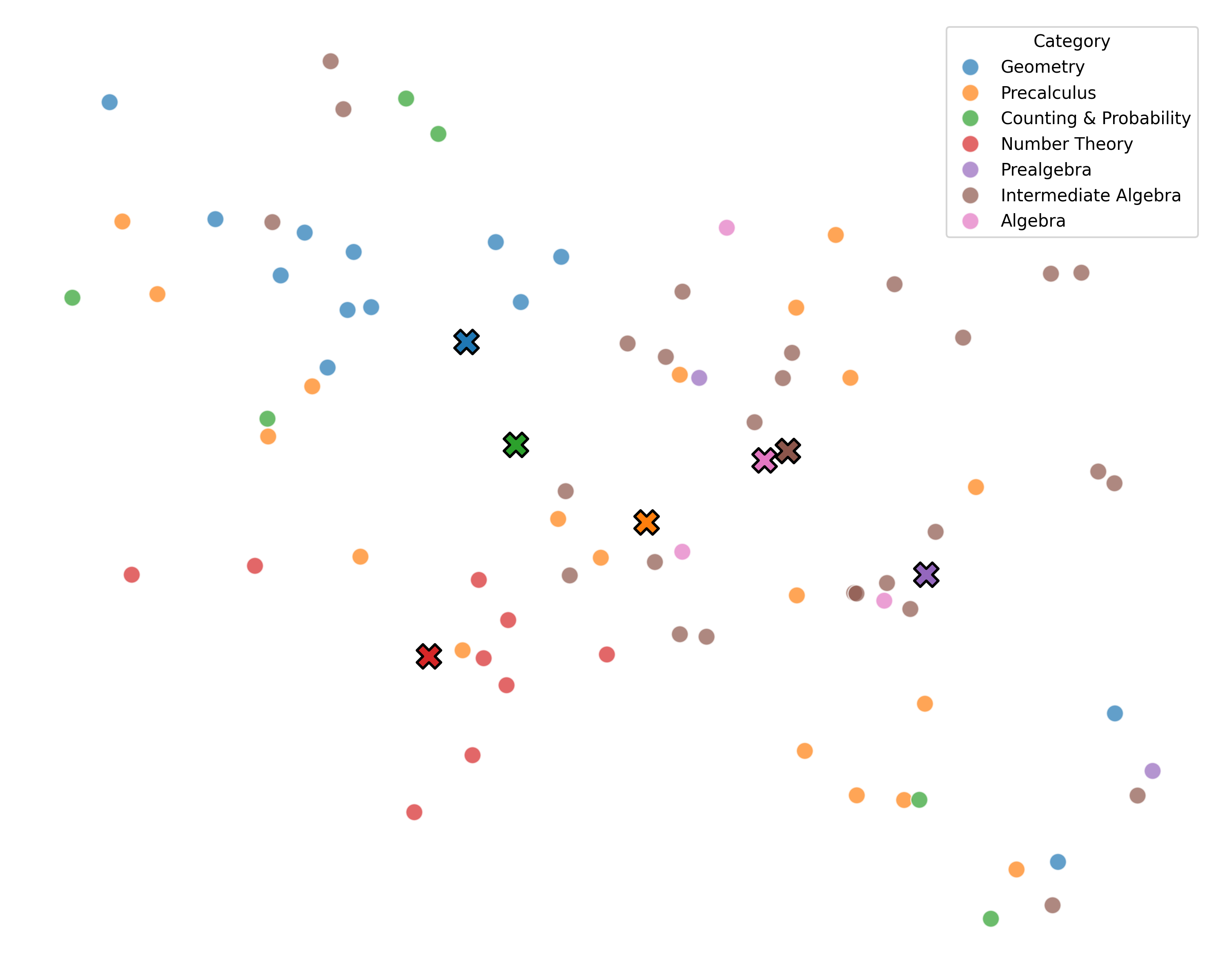} 
        \caption{}
        \label{fig:iplan42d}
    \end{subfigure}
    \caption{Illustration of the encodings of latent plans from different reasoning steps (1, 2, 3, and 4) in 2D space. We follow the same procedure as in Figure~\ref{fig:iplanrelations} and visualize the latent plan encodings using t-SNE.}
    \label{fig:iplans2d}
    \vspace{-4mm}
\end{figure*}

\begin{figure*}[t]
    \centering

    \begin{subfigure}[b]{\textwidth}
        \includegraphics[width=\linewidth]{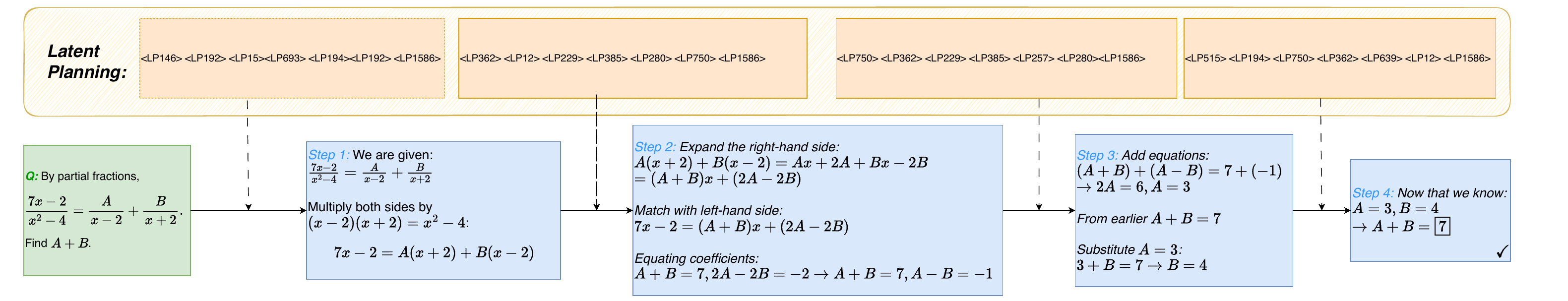} 
    \end{subfigure}
    \caption{Illustration of the reasoning process of LLMs with latent planning. In each of the four reasoning steps, Qwen2.5-7B with \emph{iCLP} first plans in the latent space, which then guides the generation of the next reasoning step.}
    \label{fig:latentdemo}
    \vspace{-4mm}
\end{figure*}

\begin{table}[tbp]
    \footnotesize
    \centering
    \caption{Comparison of different dimensions ($d_h$ and sizes (K)) of the coodbook \(\gH\).}
    \begin{adjustbox}{width=\columnwidth}
    \begin{tabular}{c|c|c|c|c|c|c}
    \toprule
    Qwen2.5               & $d_h$ & $K$ & \texttt{M}    & \texttt{CC}   & \texttt{M}$\rightarrow$\texttt{AM} & \texttt{CC}$\rightarrow$\texttt{HE} \\ \hline\hline 
    \multirow{5}{*}{3B} & 256    & 1024   & 53.7 & 14.5 & 0                  & 47.6  \\ \cline{2-2}
                        & 512    & 1024   & 56.9 & 17   & 10                 & 49.4  \\ \cline{2-2} 
                        & 1024   & 1024   & 57.2 & 17.6 & 10                 & 51.2  \\ \cline{2-2} 
                        & 512    & 2048   & 60.1 & 18.8 & 10                 & 54    \\ \cline{2-2} 
                        & 12   & 4096   & 60.1 & 19.4 & 10                 & 54.3 \\
    \bottomrule
    \end{tabular}
    \end{adjustbox}
    \label{tab:albation}
    \vspace{-4mm}
\end{table}
To better understand and visualize the latent space of the latent plans, we perform reasoning with Qwen2.5-7B using \emph{iCLP} on the \texttt{MATH} dataset and collect the encodings of latent plans across different reasoning steps. For each latent plan, we compute the average encodings by applying mean pooling over its token encodings. We then present the pairwise distance relationships between plans from different steps in Figure~\ref{fig:iplanrelations}, and the corresponding encoding structures in 2D space using t-SNE in Figure~\ref{fig:iplans2d}. To make it easier to understand how the LLM with \emph{iCLP} performs latent planning during reasoning, we provide a demonstration in Figure~\ref{fig:latentdemo}. It is evident that the LLM only needs to perform latent planning by generating a few special tokens before each reasoning step to achieve a reliable reasoning process. More importantly, compared to existing latent reasoning methods such as Coconut~\cite{Coconut-arxiv24}, \emph{iCLP} enables planning in the latent space while maintaining CoT reasoning in the language space, thereby guaranteeing strong interpretability, which is crucial for practical applications.

From Figure~\ref{fig:iplanrelations}, we observe two key patterns. First, for questions belonging to the same category in \texttt{MATH}, their latent plans exhibit clear similarity, as evidenced by the prominently highlighted diagonal regions in Figures~\ref{fig:iplan1relation},\ref{fig:iplan2relation}, \ref{fig:iplan3relation}, and~\ref{fig:iplan4relation}. This indicates that the LLM with \emph{iCLP} tends to generate similar latent plans when solving problems within the same category. Second, in the later reasoning steps—particularly in Figures~\ref{fig:iplan3relation} and~\ref{fig:iplan4relation}—latent plans become more aligned across different categories, suggesting that the model increasingly draws on similar implicit plan knowledge regardless of the specific problem type. 
From Figure~\ref{fig:iplans2d}, we observe two key patterns. First, latent plans from the same category form distinct clusters, while those from different categories are clearly separated, indicating strong category-specific structuring in the latent space. Second, the clusters from different categories are distributed around a common center, suggesting that the latent plans share a core of common implicit knowledge that generalizes well across problem types. Therefore, the result matches our motivation in Subsection~\ref{subsec:motivation}: \textbf{enabling the LLM to plan in the latent space with generalizable reasoning guidance that is reusable and transfers well across problems, thereby supporting both the generalization and accuracy of reasoning}.

\bibliography{reference}
\bibliographystyle{icml2026}


\end{document}